\documentclass[10pt,twocolumn,letterpaper]{article}

\usepackage{arxiv}              
\definecolor{cvprblue}{rgb}{0.21,0.49,0.74}
\usepackage[pagebackref,breaklinks,colorlinks,allcolors=cvprblue]{hyperref}

\def\paperID{} 
\def\confName{CVPR}
\def\confYear{2026}

\title{FactorDrive: Adaptive Multi-Step Reasoning Driven by Planning-Critical Factors for End-to-End Autonomous Driving}

\author{
    Guolei Huang\textsuperscript{1,*}
    Tengfei She\textsuperscript{2,*} 
    Yuxuan Lu\textsuperscript{3} 
    Yao Huang\textsuperscript{4}
    Yuqi Ye\textsuperscript{5}
    Yongjun Shen\textsuperscript{1,\dag} \\
    \textsuperscript{1}Southeast University \quad 
    \textsuperscript{2}Universiti Malay \quad 
    \textsuperscript{3}Ant Group \quad \\
    \textsuperscript{4}Tsinghua University \quad 
    \textsuperscript{5}Peking University
}

\usepackage{float}            
\usepackage{placeins} 
\usepackage{balance}
\usepackage{graphicx}
\usepackage{booktabs}
\usepackage{xcolor}
\usepackage{pifont}
\usepackage{multirow}
\usepackage[table]{xcolor} 
\usepackage{colortbl}
\usepackage[normalem]{ulem} 
\usepackage{booktabs,tabularx,array,ragged2e}
\usepackage{algorithm}
\usepackage{algpseudocode}
\usepackage[accsupp]{axessibility}  

\algrenewcommand\algorithmicrequire{\textbf{Input:}}
\algrenewcommand\algorithmicensure{\textbf{Output:}}
\algrenewcommand\algorithmiccomment[1]{\hfill{\footnotesize// #1}}

\begin{document}
\renewcommand{\thefootnote}{\fnsymbol{footnote}}

\maketitle

\footnotetext[0]{
    \textsuperscript{*}Equal contribution. \quad \textsuperscript{\dag}Corresponding authors. }

\begin{abstract}
Vision-language models (VLMs) have advanced scene understanding and enabled explicit reasoning in end-to-end autonomous driving. However, existing methods insufficiently integrate spatial-physical evidence into planning reasoning, while reasoning adaptation remains coarse-grained and falls short of scene-specific planning demands. Furthermore, reasoning-path optimization for higher planning quality remains largely unexplored in autonomous-driving post-training. To address these limitations, we propose FactorDrive, an end-to-end autonomous driving framework for adaptive multi-step reasoning driven by planning-critical factors (PCFs). We first perform large-scale driving-domain instruction tuning to establish foundational driving knowledge. Building on this foundation, we construct PCF-CoT, a chain-of-thought (CoT) dataset that grounds planning reasoning in trajectory-relevant spatial-physical evidence and organizes reasoning around scene-specific PCFs, enabling the composition and depth of reasoning paths to adapt to different planning demands. We further introduce Quality Search-Guided Group Relative Policy Optimization (QS-GRPO), which guides Monte Carlo Tree Search (MCTS) with trajectory-level planning rewards to discover reasoning paths with higher planning quality and uses the resulting responses to optimize the policy through GRPO, thereby improving trajectory planning performance. Extensive experiments on both open-loop (nuScenes) and closed-loop-oriented (NAVSIM) benchmarks demonstrate that FactorDrive achieves state-of-the-art planning performance.
\end{abstract}

\section{Introduction}
In recent years, autonomous driving has increasingly shifted from modular pipelines to end-to-end architectures. While modular systems offer engineering flexibility, independently optimized components can introduce information loss and error accumulation, limiting robustness and generalization in complex and long-tail scenarios~\cite{chen2024end-to-end}. End-to-end approaches mitigate these issues by jointly modeling perception, prediction, and planning within a unified framework~\cite{hu2023uniad,sun2025sparsedrive}. Nevertheless, conventional end-to-end systems are typically optimized to imitate driving patterns from limited supervision without explicitly modeling the reasoning process, restricting semantic reasoning and decision interpretability~\cite{xu2024vlmad}. Recent studies have therefore introduced vision-language models (VLMs) and vision-language-action (VLA) models into autonomous driving, leveraging large-scale pretraining and language modeling to enhance scene understanding, explicit reasoning, and action generation from multimodal observations~\cite{sima2024drivelm,zhou2026opendrivevla}. Building on these advances, chain-of-thought (CoT) reasoning has emerged as a promising approach for making planning decisions explicit in driving VLA models.

\begin{figure}[t]
    \centering
    \includegraphics[width=0.98\columnwidth]{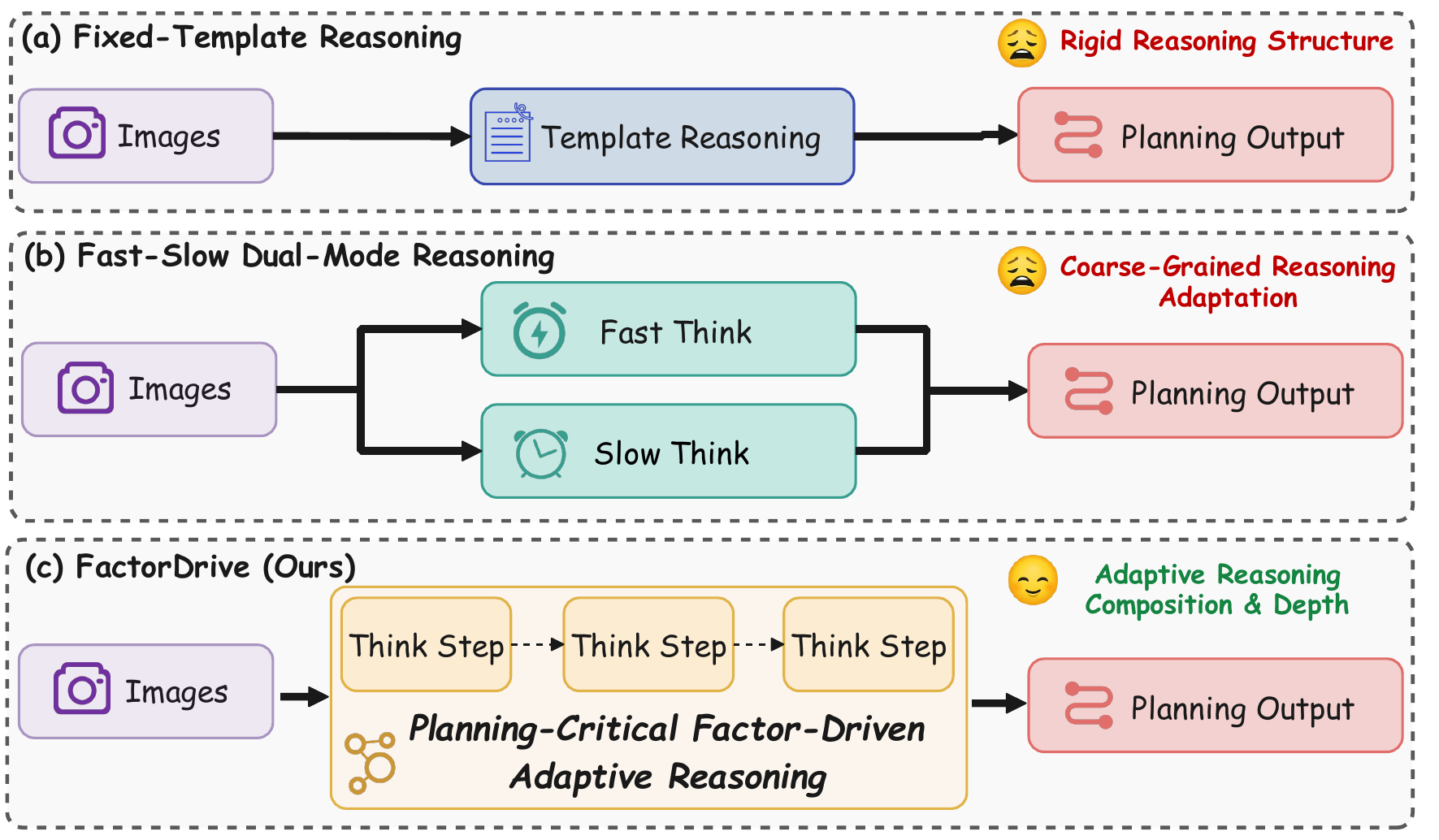}
    \caption{Comparison of reasoning paradigms for end-to-end autonomous driving. (a) Fixed-template reasoning follows a predefined reasoning chain. (b) Fast--slow dual-mode reasoning selects between two preset reasoning modes. (c) FactorDrive adaptively constructs multi-step reasoning paths according to scene-specific planning-critical factors.}
    \label{fig:introduction}
    \vspace{-6pt}
\end{figure}

Existing CoT-based methods primarily follow two reasoning paradigms. One line adopts fixed-template reasoning, organizing CoT through predefined structures such as perception--prediction--planning chains~\cite{ye2026autodrivetp3}, multi-stage logical chains with self-reflection~\cite{yuan2026autodriver-r2}, and counterfactual reasoning over planned meta-actions~\cite{peng2026counterfactual}, as illustrated in Fig.~\ref{fig:introduction}(a). The other employs fast--slow dual-mode reasoning, selecting between trajectory-only generation and CoT-enhanced planning according to scene complexity~\cite{zhou2026autovla,luo2026adathinkdrive}, as shown in Fig.~\ref{fig:introduction}(b). Together, these studies demonstrate the potential of explicit multi-step reasoning to improve planning transparency and trajectory quality.

Despite the promise of explicit reasoning in autonomous driving, existing CoT-based methods still face three fundamental limitations. \textbf{1) Spatial-physical evidence remains insufficiently integrated into planning reasoning.} Reliable trajectory planning requires precise information about lane geometry, drivable-area boundaries, traffic-control constraints, and relative agent motion. However, existing VLA methods often remain at the level of high-level semantic descriptions and fail to convert fine-grained geometric and motion constraints into planning evidence for trajectory generation~\cite{tian2025nuscenes-spatial}. \textbf{2) Reasoning adaptation remains coarse-grained.} Although driving scenes contain abundant observations, near-term decisions are primarily determined by a compact subset of planning-critical factors (PCFs), such as road structure, ego motion, and critical interacting agents, while many other observations are irrelevant or redundant to the current maneuver~\cite{renz2022plant,li2026sgdrive}. As different driving scenarios involve different PCFs and varying reasoning demands, reasoning paths should adapt their composition and depth accordingly to avoid insufficient analysis in complex scenes or superfluous reasoning in simple ones~\cite{shen2025dast}. However, fixed-template methods rely on rigid reasoning structures, while fast--slow methods merely switch between trajectory-only generation and a preset CoT mode. Neither paradigm supports such fine-grained adaptation to scene-specific planning demands. \textbf{3) Optimization of reasoning paths for higher planning quality remains under-explored in autonomous-driving post-training.} Existing GRPO-based training methods primarily optimize responses obtained through direct policy rollouts~\cite{li2026driver1,luo2026adathinkdrive}. Such sparse exploration covers only a limited portion of the reasoning space and may overlook reasoning paths that yield better planning outcomes~\cite{wu2026deepsearch}. Systematically discovering and optimizing such reasoning paths therefore remains a key challenge for improving trajectory planning performance.

To address these limitations, we propose FactorDrive, an end-to-end autonomous driving framework for adaptive multi-step reasoning driven by planning-critical factors, as illustrated in Fig.~\ref{fig:introduction}(c). FactorDrive adopts a two-stage supervised fine-tuning (SFT) strategy. In the first stage, we perform driving-domain instruction tuning on a large-scale driving corpus to adapt the base VLM to autonomous driving and establish foundational driving knowledge. In the second stage, we further fine-tune the model on PCF-CoT, a chain-of-thought dataset organized around PCFs that grounds planning reasoning in spatial-physical evidence relevant to trajectory generation and enables the composition and depth of reasoning paths to adapt to scene-specific planning demands. To further improve trajectory planning through explicit reasoning-path optimization, we introduce Quality Search-Guided Group Relative Policy Optimization (QS-GRPO), which guides Monte Carlo Tree Search (MCTS)~\cite{silver2017mcts} with trajectory-level planning rewards to discover reasoning paths with higher planning quality and optimizes the policy through GRPO~\cite{guo2025deepseek-r1} using the complete responses generated along these paths. We evaluate FactorDrive on the open-loop nuScenes~\cite{caesar2020nuscenes} and closed-loop-oriented NAVSIM~\cite{dauner2024navsim} benchmarks, where it achieves superior planning performance across both settings.

Our main contributions are summarized as follows:
\begin{itemize}
    \item We propose FactorDrive, an end-to-end autonomous driving framework for adaptive multi-step reasoning driven by planning-critical factors. We further construct PCF-CoT, a chain-of-thought dataset that organizes adaptive reasoning around planning-critical factors and integrates planning-relevant spatial-physical evidence into supervision for future trajectory generation.
    \item We introduce QS-GRPO, a quality-guided post-training method, which first utilizes MCTS with trajectory-level planning rewards to discover reasoning paths with higher planning quality and then uses the resulting responses to optimize the policy through GRPO, thereby improving the driving planning performance.
    \item Extensive experiments demonstrate that FactorDrive achieves new state-of-the-art performance on both the nuScenes and NAVSIM benchmarks, validating the effectiveness and generalizability of the proposed approach.
\end{itemize}

\begin{figure*}[t]
    \centering
    \includegraphics[width=0.9\textwidth]{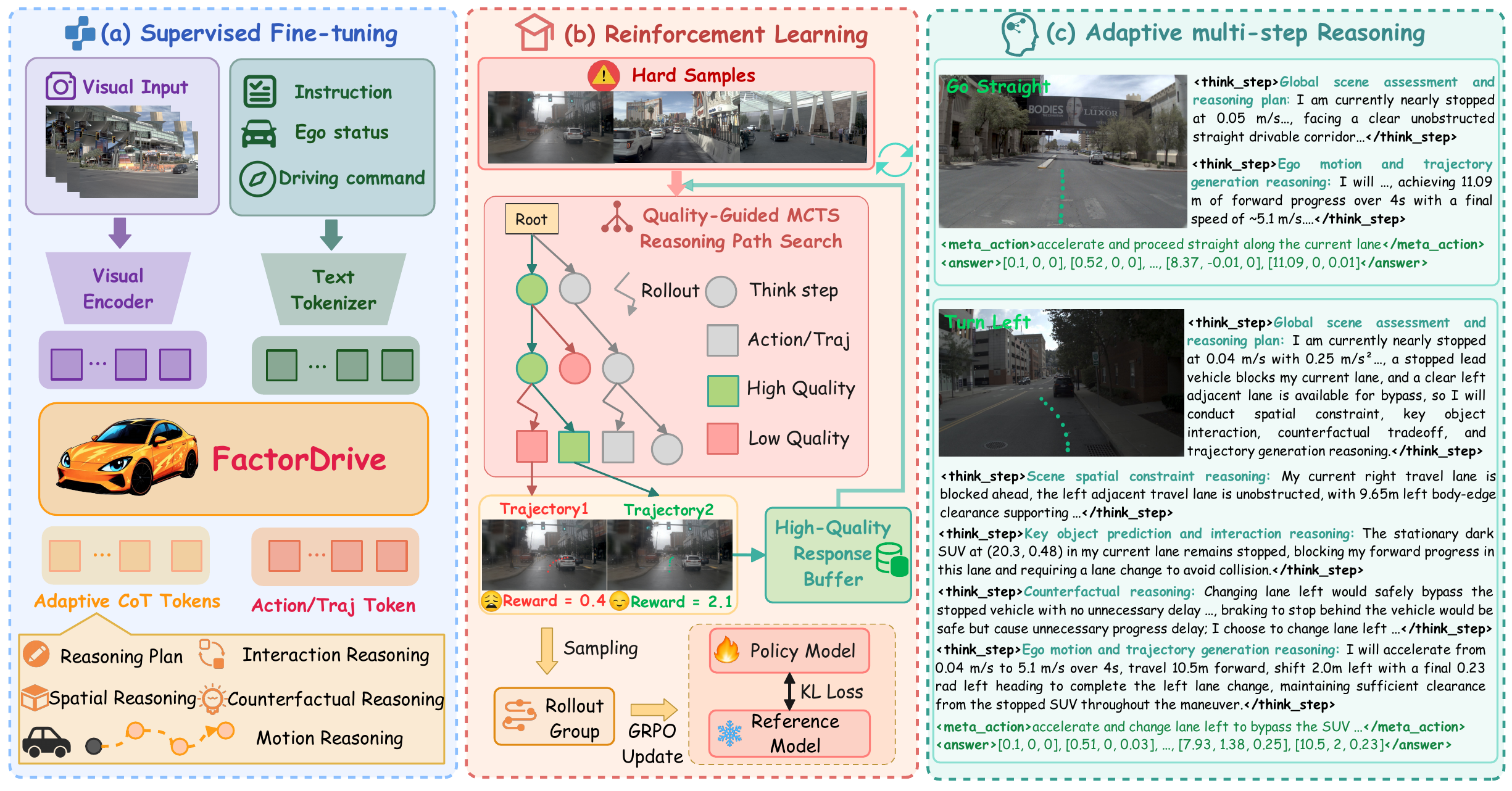}
    \caption{
    Overview of FactorDrive. (a) Adaptive CoT reasoning and trajectory generation capability is achieved through supervised fine-tuning. (b) Planning-quality optimization explores and reinforces better reasoning paths via reinforcement learning. (c) FactorDrive performs scene-adaptive multi-step reasoning over PCFs for decision and trajectory generation.
    }
    \label{fig:workflow}
    \vspace{-4pt}
\end{figure*}

\section{Related Work}
\subsection{End-to-End Autonomous Driving Methods}
End-to-end autonomous driving replaces modular pipelines with unified models such as ST-P3 \cite{hu2022stp3}, UniAD \cite{hu2023uniad}, and VAD \cite{jiang2023vad}, which jointly optimize perception, prediction, and planning yet offer limited explicit modeling of the high-level semantic reasoning underlying planning decisions \cite{xu2024vlmad}. Recent VLM-based approaches, including DriveVLM \cite{tian2025drivevlm}, EMMA \cite{hwang2025emma}, and OmniDrive \cite{wang2025omnidrive}, introduce language modeling for scene understanding, decision-making, and trajectory generation. To improve interpretability, CoT-based methods adopt predefined reasoning structures, as in $AutoDrive\text{-}P^3$ \cite{ye2026autodrivetp3} and DriveCoT \cite{wang2024drivecot}, or fast--slow reasoning mechanisms, as in AutoVLA \cite{zhou2026autovla} and AdaThinkDrive \cite{luo2026adathinkdrive}. Nevertheless, existing approaches neither adequately integrate spatial-physical evidence into planning reasoning nor support fine-grained adaptation to scene-specific planning demands. In contrast, FactorDrive grounds planning reasoning in spatial-physical evidence and adapts reasoning-path composition and depth according to scene-specific PCFs.

\subsection{Reinforcement Learning for Post-Training}
Reinforcement learning (RL) has become a key paradigm for improving reasoning during large language model post-training. PPO \cite{schulman2017ppo} enables stable policy optimization for RLHF, DPO \cite{rafailov2023dpo} simplifies preference learning through direct optimization, and GRPO \cite{shao2024deepseekmath} replaces value-function estimation with group-relative advantages, reducing computational overhead. Large-scale RL further demonstrates its potential to elicit advanced reasoning capabilities \cite{guo2025deepseek-r1}. In autonomous driving, recent VLA methods adopt GRPO with task-specific rewards for physical consistency~\cite{yuan2026autodriver-r2}, multi-task learning~\cite{ye2026autodrivetp3}, adaptive reasoning~\cite{luo2026adathinkdrive}, and planning safety~\cite{tang2026planr1}. However, these methods largely rely on direct policy rollouts, limiting the exploration of reasoning paths with higher planning quality. To address this limitation, we propose QS-GRPO, which utilizes MCTS with trajectory-level planning rewards to discover reasoning paths with higher planning quality for subsequent training through GRPO.

\section{Method}
\subsection{Overview}
\label{sec:overview}
FactorDrive is a VLM-based end-to-end trajectory planning framework that performs adaptive multi-step reasoning driven by planning-critical factors (PCFs) to predict the future ego trajectory from multimodal driving inputs. Given the input $z=(q,\mathbf{V},\mathbf{e},c_{\mathrm{nav}})$, where $q$ denotes the task instruction, $\mathbf{V}$ comprises a sequence of four front-view camera frames sampled at $2\,\mathrm{Hz}$ to capture short-term visual dynamics, $\mathbf{e}$ includes the current velocity, acceleration, and historical ego trajectory, and $c_{\mathrm{nav}}$ denotes the high-level navigation command, including Go Straight, Turn Left, and Turn Right, the model generates a structured response $Y=(T,m,\hat{\mathbf{P}})$, where $T$ denotes the reasoning path driven by planning-critical factors, $m$ represents the high-level meta-action, and $\hat{\mathbf{P}}=\{\hat{\mathbf{p}}_t\}_{t=1}^{H}$ denotes the predicted ego trajectory over the next $4\,\mathrm{s}$. Each waypoint is defined as $\hat{\mathbf{p}}_t=(\hat{x}_t,\hat{y}_t,\hat{\psi}_t)$, where $\hat{x}_t$, $\hat{y}_t$, and $\hat{\psi}_t$ denote its position and heading angle in the ego-centric coordinate frame.

\begin{figure*}[t]
    \centering
    \includegraphics[width=0.98\textwidth]{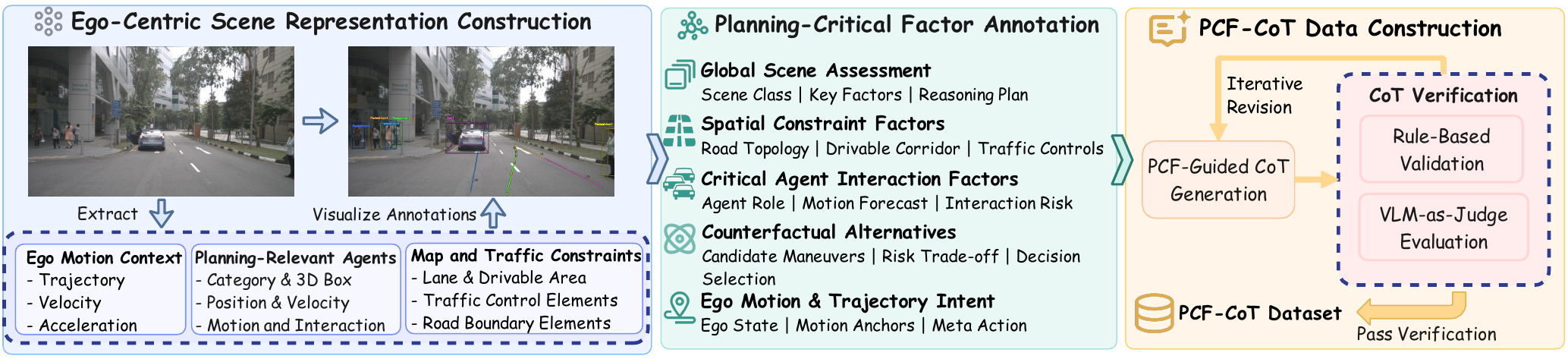}
    \caption{
    PCF-CoT construction pipeline. It comprises ego-centric scene representation, planning-critical factor annotation, and PCF-guided CoT generation with iterative verification.
    }
    \label{fig:PCF-COT}
\end{figure*}

Specifically, we first construct a Driving-Domain Instruction Corpus and the PCF-CoT dataset to provide supervision for driving-domain knowledge acquisition and adaptive multi-step reasoning driven by planning-critical factors, respectively. FactorDrive is then trained through two-stage supervised fine-tuning, followed by QS-GRPO post-training, which optimizes reasoning paths using trajectory-level rewards to further improve planning performance. The overall framework is illustrated in Fig. \ref{fig:workflow}.

\subsection{Data Construction}
\subsubsection{Driving-Domain Instruction Corpus.}
To bridge the domain gap between general-purpose VLMs and autonomous driving, we first construct a large-scale Driving-Domain Instruction Corpus to establish foundational driving-domain capabilities. Specifically, we collect and curate approximately one million driving-scene question–answer pairs from publicly available datasets, including DriveLM~\cite{sima2024drivelm}, LingoQA~\cite{marcu2024lingoqa}, OmniDrive~\cite{wang2025omnidrive}, DriveLMM-o1~\cite{ishaq2025drivelmm-o1}, nuScenes-QA~\cite{qian2024nuscenes-qa}, Impromptu-VLA~\cite{chi2026impromptu-vla}, and STRIDE-QA~\cite{ishihara2026stride-qa}. The resulting corpus covers a diverse range of tasks, including perceptual understanding, agent motion prediction, trajectory planning, traffic-rule reasoning, counterfactual reasoning, and spatiotemporal physical relation modeling.

\subsubsection{PCF-CoT Dataset.}
To integrate planning-relevant spatial-physical evidence into planning reasoning and enable adaptive multi-step reasoning, we construct PCF-CoT, a chain-of-thought dataset organized around planning-critical factors. PCFs refer to the subset of scene and motion factors that substantially constrain future ego planning, including road geometry, lane structure and drivable areas, traffic-control elements, critical interacting agents, and ego-motion history. Rather than exhaustively describing all visible elements, PCF-CoT identifies the spatial-physical evidence associated with these factors, such as geometric constraints, relative motion, interaction risks, and traffic rules, and structures the reasoning process around this evidence to support future trajectory generation.

As illustrated in Fig.~\ref{fig:PCF-COT}, PCF-CoT is constructed through a three-stage annotation pipeline. First, we integrate historical video frames, ego-motion states, navigation commands, planning-relevant agents, map information, and traffic constraints, and transform positions, velocities, and trajectories into a unified ego-centric coordinate system. We further visualize critical agents, lane structures, drivable areas, road boundaries, and traffic-control elements to construct a compact ego-centric scene representation that preserves planning-relevant spatial geometry, relative motion, and traffic constraints. Second, we employ Seed-2.0-Pro~\cite{bytedanceseed2026seed20} to annotate planning-critical factors in a structured manner, converting raw visual, map, and motion information into planning-relevant spatial-physical evidence, including global scene assessment, spatial constraints, interactions with critical agents, counterfactual action candidates, and ego-motion and trajectory intent. Finally, conditioned on the annotated PCFs, Seed-2.0-Pro generates adaptive CoT reasoning. It first produces a Reasoning Plan that identifies the active PCFs and determines the reasoning requirements of the current scene. The model then adaptively selects and organizes the required units from Spatial Reasoning, Interaction Reasoning, Counterfactual Reasoning, and Motion Reasoning, allowing reasoning-path composition and depth to vary with scene-specific planning demands. To ensure annotation quality, we combine rule-based format validation with VLM-as-Judge evaluation of planning relevance, spatial-physical plausibility, and output consistency. To reduce self-evaluation bias~\cite{panickssery2024llm}, we use Qwen3.6-Plus~\cite{qwenteam2026qwen36plus} as an independent judge. Samples that fail the evaluation are iteratively revised based on the judge’s feedback, while those that remain invalid after multiple refinement rounds are discarded.

Through the above annotation pipeline, we construct PCF-CoT, a high-quality adaptive multi-step CoT dataset organized around planning-critical factors. It includes the nuScenes-PCF subset, with 14,802 samples from 700 nuScenes training scenes, and the NAVSIM-PCF subset, with 102,623 samples from 1,192 NAVSIM training scenes.

\subsection{Two-Stage Supervised Fine-Tuning}
To adapt a general-purpose VLM to autonomous driving and equip it with adaptive multi-step reasoning capabilities, we employ a two-stage supervised fine-tuning strategy. In the first stage, we perform driving-domain instruction tuning on the collected Driving-Domain Instruction Corpus, enabling the model to acquire fundamental driving knowledge from diverse QA supervision, including scene understanding, traffic semantics, motion prediction, planning knowledge, and spatiotemporal physical reasoning.
In the second stage, we further fine-tune the model on the proposed PCF-CoT dataset to learn reasoning driven by planning-critical factors. As introduced in the overview, for each multimodal driving input $z$, the structured target response $Y$ is serialized into a token sequence $\mathbf{y}=(y_1,\ldots,y_N)$ comprising the reasoning path, meta-action, and future trajectory. The model is optimized using the autoregressive negative log-likelihood objective:
\begin{equation}
\mathcal{L}_{\mathrm{SFT}}=-\frac{1}{N}\sum_{i=1}^{N}\log p_{\theta}\!\left(y_i \mid y_{<i}, z\right),
\label{eq:sft_objective}
\end{equation}
where $y_i$ denotes the $i$-th target token, $y_{<i}$ denotes preceding target tokens, and $\theta$ denotes the model parameters. This stage jointly optimizes planning-critical reasoning and trajectory generation in an autoregressive framework.

\subsection{QS-GRPO Post-Training}
After SFT, the model is equipped with adaptive multi-step reasoning capabilities. To further improve trajectory planning through explicit reasoning-path optimization, we propose QS-GRPO, a quality-guided post-training method that explores and reinforces reasoning paths yielding better planning outcomes.

\subsubsection{Quality-Guided MCTS for Reasoning Path Search.}
Inspired by the search-augmented reasoning approaches of DeepSearch~\cite{wu2026deepsearch} and STAIR~\cite{zhang2025stair}, QS-GRPO incorporates quality-guided MCTS to search for reasoning paths that achieve higher planning rewards than direct autoregressive sampling. To improve search efficiency, we construct a hard-sample subset $\mathcal{D}_{\mathrm{hard}}$ based on the planning performance of the SFT model and apply MCTS only to $\xi\in\mathcal{D}_{\mathrm{hard}}$. This concentrates the limited search budget on complex scenarios that more readily expose model failure modes and offer greater optimization potential.

For each hard sample $\xi$, let $z_\xi$ denote its multimodal driving input. Conditioned on $z_\xi$, MCTS constructs a reasoning-state tree, where each node represents a partial reasoning state and each edge corresponds to a candidate reasoning step proposed by the old policy. A complete response from the root to a terminal node is denoted by $\tau$, comprising reasoning over planning-critical factors, a meta-action, and a future trajectory. Each response is evaluated using a trajectory-level planning reward, which is backpropagated through the search tree to allocate more search budget to branches that yield higher-quality planning outcomes.

\noindent{\textbf{Quality-Guided Search.}}
MCTS proceeds through selection, expansion, rollout, and backpropagation. During selection, we adopt the PUCT criterion (Silver et al. 2017), which balances the estimated value of each reasoning branch with its prior probability under the old policy. At a selected non-terminal leaf, the old policy proposes multiple candidate reasoning steps to expand the tree. Rollouts from the expanded nodes generate complete responses, each scored by
\begin{equation}
r(\tau, \xi)
=
\lambda_{\mathrm{fmt}} R_{\mathrm{fmt}}
+
\lambda_{\mathrm{traj}} R_{\mathrm{traj}},
\label{eq:reward}
\end{equation}
where $R_{\mathrm{fmt}}$ measures output-format correctness, $R_{\mathrm{traj}}$ evaluates the planning quality of the predicted trajectory, and $\lambda_{\mathrm{fmt}}$ and $\lambda_{\mathrm{traj}}$ are the weighting coefficients for the two reward terms. The reward is backpropagated to update the statistics of the visited branches, progressively directing the search toward reasoning paths associated with higher-quality planning outcomes.

\noindent{\textbf{High-Quality Response Buffer.}} To reduce redundant search and stabilize candidate-group construction, we maintain a High-Quality Response Buffer for each hard sample. The buffer retains only responses whose rewards exceed a predefined quality threshold, thereby preserving reliable search results across training iterations.

Specifically, if sample $\xi$ has no cached response, i.e., $\mathcal{B}(\xi)=\varnothing$, we perform a full MCTS search and denote the set of candidate responses obtained during the search by $\Omega_{\mathrm{mcts}}(\xi)$. We then select the highest-reward response as $\tau_{\mathrm{mcts}}^{*}=\arg\max_{\tau\in\Omega_{\mathrm{mcts}}(\xi)}r(\tau,\xi)$. This response is stored in the buffer only if $r(\tau_{\mathrm{mcts}}^{*},\xi)\geq\delta_{\mathrm{hq}}$, i.e., $\mathcal{B}(\xi)\leftarrow\tau_{\mathrm{mcts}}^{*}$, where $\delta_{\mathrm{hq}}$ denotes the high-quality reward threshold.

If sample $\xi$ already has a cached response, subsequent training iterations reuse $\mathcal{B}(\xi)$ without repeating the full MCTS search. Instead, we perform direct rollouts following standard GRPO sampling to obtain the candidate set $\Omega_{\mathrm{roll}}(\xi)$. We denote the highest-reward rollout response by $\tau_{\mathrm{roll}}^{*}=\arg\max_{\tau\in\Omega_{\mathrm{roll}}(\xi)}r(\tau,\xi)$. If $r(\tau_{\mathrm{roll}}^{*},\xi)>r(\mathcal{B}(\xi),\xi)$, the cached response is updated as $\mathcal{B}(\xi)\leftarrow\tau_{\mathrm{roll}}^{*}$. This mechanism reserves costly MCTS searches for samples that have not yet yielded reliable high-quality responses, thereby concentrating the search budget on hard scenarios with greater remaining optimization potential.

\noindent{\textbf{Confidence-Aware Group Construction.}} After completing MCTS or direct rollout, we construct a candidate response pool for QS-GRPO optimization. These two procedures correspond to different buffer states at the start of the current training iteration and therefore require different candidate-pool construction strategies. For a sample without a previously cached response, we combine the highest-reward response found by MCTS with candidate responses directly rolled out from the root node. If a cached response already exists, we form the candidate response pool by combining it with newly generated direct-rollout candidates:
\begin{equation}
\Omega(\xi)= \begin{cases} \{\tau_{\mathrm{mcts}}^{*}\}\cup\Omega_{\mathrm{root}}(\xi), & \mathcal{B}(\xi)=\varnothing, \\ \{\mathcal{B}(\xi)\}\cup\Omega_{\mathrm{roll}}(\xi), & \mathcal{B}(\xi)\neq\varnothing, \end{cases}
\label{eq:candidate}
\end{equation}
where $\Omega_{\mathrm{root}}(\xi)$ denotes the candidate responses directly sampled from the root node by the old policy, and $\Omega_{\mathrm{roll}}(\xi)$ denotes the direct-rollout candidates generated when a cached response is available. We retain the highest-reward response in $\Omega(\xi)$, denoted by $\tau^*$, ensuring that each training group retains its highest-reward planning candidate. The remaining candidates are selected according to the confidence assigned by the old policy. Given a candidate response $\tau=(y_1,\ldots,y_{|\tau|})$, its confidence is defined as the length-normalized log-likelihood:
\begin{equation}
C(\tau\mid z_\xi)=\frac{1}{|\tau|}\sum_{t=1}^{|\tau|}\log\pi_{\theta_{\mathrm{old}}}\left(y_t\mid y_{<t},z_\xi\right),
\label{eq:confidence}
\end{equation}
where $y_t$ denotes the $t$-th generated token and $\pi_{\theta_{\mathrm{old}}}$ is the old policy used to generate the candidate responses. The final training group is defined as
\begin{equation}
\mathcal{G}(\xi)=\{\tau^*\}\cup\mathrm{TopConf}_{G-1}\left(\Omega(\xi)\setminus\{\tau^*\}\right),
\label{eq:group}
\end{equation}
where $\operatorname{TopConf}_{G-1}(\cdot)$ selects the $G-1$ remaining responses with the highest confidence scores. This construction preserves the best response while retaining candidates that the old policy is most likely to generate. High-confidence but low-reward responses expose the model’s overconfident failure modes and receive lower group-relative advantages, reducing their likelihood during GRPO optimization.

\begin{table*}[t]
\centering
\small
\setlength{\tabcolsep}{12pt}
\setlength{\arrayrulewidth}{0.4pt}

\begin{tabularx}{\textwidth}{l|c|*{4}{c}|*{4}{c}}
\toprule

\multirow{2}{*}{\textbf{Method}}
&
\multirow{2}{*}{\textbf{Venue}}
&
\multicolumn{4}{c|}{\textbf{L2 (m)} $\downarrow$}
&
\multicolumn{4}{c}{\textbf{Collision (\%)} $\downarrow$}
\\

\cmidrule(lr){3-6}
\cmidrule(lr){7-10}

&

&
1s
&
2s
&
3s
&
Avg.
&
1s
&
2s
&
3s
&
Avg.
\\

\midrule
\multicolumn{10}{c}{\textit{Non-Autoregressive Methods}}
\\
\midrule

ST-P3~\cite{hu2022stp3}
& ECCV 2022
& 1.33 & 2.11 & 2.90 & 2.11
& 0.23 & 0.62 & 1.27 & 0.71
\\

UniAD~\cite{hu2023uniad}
& CVPR 2023
& 0.44 & 0.67 & 0.96 & 0.69
& 0.04 & 0.08 & 0.23 & 0.12
\\

VAD~\cite{jiang2023vad}
& ICCV 2023
& 0.17 & 0.34 & 0.60 & 0.37
& 0.07 & 0.10 & 0.24 & 0.14
\\

\midrule
\multicolumn{10}{c}{\textit{Autoregressive Methods}}
\\
\midrule


DriveVLM~\cite{tian2025drivevlm}
& CoRL 2025
& 0.18 & 0.34 & 0.68 & 0.40
& 0.10 & 0.22 & 0.45 & 0.27
\\

EMMA~\cite{hwang2025emma}
& TMLR 2025
& \underline{0.14} & 0.29 & 0.54 & 0.32
& -- & -- & -- & --
\\

RDA-Driver~\cite{huang2024rdadriver}
& ECCV 2024
& 0.17 & 0.37 & 0.69 & 0.40
& \underline{0.01} & \underline{0.05} & 0.26 & 0.10
\\

OmniDrive~\cite{wang2025omnidrive}
& CVPR 2025
& \underline{0.14} & 0.29 & 0.55 & 0.33
& \textbf{0.00} & 0.13 & 0.78 & 0.30
\\

AutoVLA~\cite{zhou2026autovla}
& NeurIPS 2025
& 0.21 & 0.38 & 0.60 & 0.40
& 0.13 & 0.18 & 0.28 & 0.20
\\

OpenDriveVLA~\cite{zhou2026opendrivevla}
& AAAI 2026
& 0.15 & 0.31 & 0.55 & 0.33
& \underline{0.01} & 0.08 & 0.21 & 0.10
\\

Drive-R1~\cite{li2026driver1}
& AAAI 2026
& \underline{0.14} & \underline{0.28} & \textbf{0.50} & \underline{0.31}
& 0.02 & 0.06 & \underline{0.19} & 0.09
\\

$AutoDrive\text{-}P^3$~\cite{ye2026autodrivetp3}
& ICLR 2026
& 0.15 & 0.30 & 0.54 & 0.33
& \textbf{0.00} & \textbf{0.02}
& \textbf{0.15} & \textbf{0.06}
\\

\specialrule{\lightrulewidth}{0pt}{0pt}

\textbf{FactorDrive (Ours)}
& \textbf{--}
& \textbf{0.12}
& \textbf{0.26}
& \underline{0.52}
& \textbf{0.30}
& \textbf{0.00}
& \underline{0.05}
& \underline{0.19}
& \underline{0.08}
\\

\specialrule{\heavyrulewidth}{0pt}{0pt}
\end{tabularx}

\caption{Open-loop planning performance comparison on nuScenes.}
\label{tab:nuscenes_results}
\end{table*}

\begin{table*}[t]
\centering
\small
\setlength{\tabcolsep}{8.5pt}
\setlength{\arrayrulewidth}{0.4pt}
\renewcommand{\arraystretch}{1.05}

\begin{tabularx}{\textwidth}{
l|
c|
*{2}{c}|
*{5}{c}|
c
}
\toprule

\textbf{Method}
&
\textbf{Venue}
&
\textbf{Image}
&
\textbf{LiDAR}
&
\textbf{NC} $\uparrow$
&
\textbf{DAC} $\uparrow$
&
\textbf{EP} $\uparrow$
&
\textbf{TTC} $\uparrow$
&
\textbf{Comf.} $\uparrow$
&
\textbf{PDMS} $\uparrow$
\\

\midrule



UniAD~\cite{hu2023uniad}
& CVPR 2023
& $\checkmark$
& $\times$
& 97.8
& 91.9
& 78.8
& 92.9
& \textbf{100.0}
& 83.4
\\


TransFuser~\cite{chitta2022transfuser}
& TPAMI 2022
& $\checkmark$
& $\checkmark$
& 97.7
& 92.8
& 79.2
& 92.8
& \textbf{100.0}
& 84.0
\\

PARA-Drive~\cite{Weng2024paradrive}
& CVPR 2024
& $\checkmark$
& $\times$
& 97.9
& 92.4
& 79.3
& 93.0
& 99.8
& 84.0
\\



DiffusionDrive~\cite{liao2025diffusiondrive}
& CVPR 2025
& $\checkmark$
& $\checkmark$
& 98.2
& 96.2
& 82.2
& 94.7
& \textbf{100.0}
& 88.1
\\

WoTE~\cite{li2025wote}
& ICCV 2025
& $\checkmark$
& $\checkmark$
& 98.5
& 96.8
& 81.9
& 94.9
& \underline{99.9}
& 88.3
\\

AutoVLA~\cite{zhou2026autovla}
& NeurIPS 2025
& $\checkmark$
& $\times$
& 98.4
& 95.6
& 81.9
& \textbf{98.0}
& \underline{99.9}
& 89.1
\\

AdaThinkDrive~\cite{luo2026adathinkdrive}
& ICRA 2026
& $\checkmark$
& $\times$
& 98.4
& \underline{97.8}
& \underline{84.4}
& 95.2
& \textbf{100.0}
& 90.3
\\

$AutoDrive\text{-}P^3$~\cite{ye2026autodrivetp3}
& ICLR 2026
& $\checkmark$
& $\times$
& \underline{99.1}
& 97.4
& \textbf{84.8}
& 96.5
& \textbf{100.0}
& 90.6
\\

\specialrule{\lightrulewidth}{0pt}{0pt}

\textbf{FactorDrive (Ours)}
& --
& $\checkmark$
& $\times$
& \textbf{99.3}
& \textbf{98.2}
& 83.6
& \underline{97.6}
& \textbf{100.0}
& \textbf{91.0}
\\

\specialrule{\heavyrulewidth}{0pt}{0pt}

\end{tabularx}

\caption{
Planning performance comparison on NAVSIM using closed-loop-oriented metrics.
}
\label{tab:navsim_results}
\end{table*}

\subsubsection{QS-GRPO Optimization Objective.}
Given the confidence-aware training group above, QS-GRPO optimizes the policy using group-relative advantages. For each hard sample $\xi$, we obtain a group $\mathcal{G}(\xi)=\{\tau_i\}_{i=1}^{G}$, where each response $\tau_i$ receives reward $r_i=r(\tau_i,\xi)$. Following the standard GRPO formulation~\cite{guo2025deepseek-r1}, the group-relative advantage is computed as
\begin{equation}
\hat{A}_i=\frac{r_i-\operatorname{mean}(\{r_j\}_{j=1}^G)}{\operatorname{std}(\{r_j\}_{j=1}^G)+\delta},
\label{eq:advantage}
\end{equation}
where $\delta$ is a small constant for numerical stability. Using these advantages, the training objective is defined as
\begin{align}
\mathcal{J}(\theta)
={}&
\mathbb{E}_{\xi,\mathcal{G}(\xi)}
\Biggl[
\frac{1}{G}\sum_{i=1}^{G}
\min\!\Bigl(
\rho_i(\theta)\hat{A}_i,\,
\bar{\rho}_i(\theta)\hat{A}_i
\Bigr)
\notag
\\
&\qquad
-\beta D_{\mathrm{KL}}
\!\left(
\pi_\theta \,\|\, \pi_{\mathrm{ref}}
\right)
\Biggr],
\label{eq:qs_grpo_objective}
\\
\bar{\rho}_i(\theta)
={}&
\operatorname{clip}
\!\left(
\rho_i(\theta),\,1-\epsilon,\,1+\epsilon
\right),
\label{eq:clipped_ratio}
\end{align}
where $\rho_i(\theta)=\frac{\pi_{\theta}(\tau_i\mid z_\xi)}{\pi_{\theta_{\mathrm{old}}}(\tau_i\mid z_\xi)}$ denotes the importance ratio between the current and old policies, $\epsilon$ is the clipping coefficient, $\beta$ controls the KL regularization strength, and $\pi_{\mathrm{ref}}$ is the fixed SFT reference model. Unlike standard GRPO, which forms training groups solely through direct rollouts from the old policy, QS-GRPO integrates quality-guided MCTS with confidence-aware candidate selection during group construction. This enables the policy to reinforce high-reward reasoning paths while suppressing high-confidence but low-reward paths.

\section{Experiments}
\subsection{Benchmarks and Metrics}
We evaluate FactorDrive on two widely used autonomous driving benchmarks, nuScenes~\cite{caesar2020nuscenes} and NAVSIM~\cite{dauner2024navsim}, covering open-loop trajectory prediction and closed-loop-oriented planning evaluation, respectively. On nuScenes, following the ST-P3 protocol~\cite{hu2022stp3}, we report trajectory $\mathrm{L2}$ error and collision rate. On NAVSIM, we report the PDM Score (PDMS), which aggregates no-at-fault collision (NC), drivable area compliance (DAC), time-to-collision (TTC), ego progress (EP), and comfort to assess overall planning quality.

\subsection{Implementation Details}
\noindent{\textbf{Hard-Sample Selection.}} We train separate SFT models on nuScenes-PCF and NAVSIM-PCF. For QS-GRPO post-training, each SFT model is evaluated once on the full training split of the corresponding benchmark, and hard samples are selected according to planning performance. We retain nuScenes samples with $\mathrm{L2}_{3\mathrm{s}}>0.5$, yielding 6,407 samples, and NAVSIM samples with $\mathrm{PDMS}<0.80$, yielding 4,910 samples.

\noindent{\textbf{Reward Design.}}
The binary format reward $R_{\mathrm{fmt}}$ is set to $1$ if the response follows the required output format and $0$ otherwise. For nuScenes, the trajectory-quality reward combines the average trajectory error reward $R_{\mathrm{L2}}=\exp\!\left(-\mathrm{L2}(\hat{\mathbf P},\mathbf P^{\mathrm{gt}})/\alpha_{\mathrm{L2}}\right)$ and the final displacement error reward $R_{\mathrm{FDE}}=\exp\!\left(-\mathrm{FDE}(\hat{\mathbf P},\mathbf P^{\mathrm{gt}})/\alpha_{\mathrm{FDE}}\right)$, where $\mathrm{L2}(\cdot)$ is the mean Euclidean displacement error over the prediction horizon and $\mathrm{FDE}(\cdot)=\|\hat{\mathbf p}_{3\mathrm{s}}-\mathbf p^{\mathrm{gt}}_{3\mathrm{s}}\|_2$ is the endpoint error at $3\,\mathrm{s}$. We set $\alpha_{\mathrm{L2}}=\alpha_{\mathrm{FDE}}=1.0$. For NAVSIM, the trajectory-quality reward combines $R_{\mathrm{L2}}$ with a PDMS reward. Across both benchmarks, the format and trajectory-quality reward weights are set to $\lambda_{\mathrm{fmt}}:\lambda_{\mathrm{traj}}=1:5$.

\noindent{\textbf{Training Details.}} 
We adopt Qwen3-VL-8B~\cite{Qwen3-VL} as the base model. During driving-domain instruction tuning, the model is trained for 1 epoch with an effective batch size of 128 and a learning rate of $5\times10^{-6}$. We then fine-tune separate nuScenes-PCF and NAVSIM-PCF models for 3 and $4$ epochs, respectively, using an effective batch size of 16 and a learning rate of $1\times10^{-5}$. All training is conducted on eight NVIDIA A800 80GB GPUs. For QS-GRPO post-training, the policy is initialized from the corresponding SFT checkpoint, with a learning rate of $1\times10^{-6}$, a group size of $G=8$, and a KL coefficient of $\beta=0.001$. The nuScenes and NAVSIM models are trained for 2,400 and 3,065 optimization steps, respectively.

\subsection{Performance Comparison}

\noindent{\textbf{Results on nuScenes.}} As shown in Tab.~\ref{tab:nuscenes_results}, we compare FactorDrive with representative non-autoregressive and autoregressive methods on the nuScenes open-loop planning benchmark. FactorDrive achieves L2 errors of $0.12\,\mathrm{m}$, $0.26\,\mathrm{m}$, and $0.52\,\mathrm{m}$ at the $1\,\mathrm{s}$, $2\,\mathrm{s}$, and $3\,\mathrm{s}$ horizons, respectively. It delivers the best performance at $1\,\mathrm{s}$ and $2\,\mathrm{s}$, while remaining highly competitive at $3\,\mathrm{s}$. Its average L2 error further decreases to $0.30\,\mathrm{m}$, outperforming the previous best result of $0.31\,\mathrm{m}$ achieved by Drive-R1 and establishing a new state of the art (SOTA). These results demonstrate that FactorDrive improves short-term trajectory accuracy while effectively limiting error accumulation over longer horizons. In terms of safety, FactorDrive achieves an average collision rate of $0.08\%$, ranking second among methods that report this metric, behind only $AutoDrive\text{-}P^3$ at $0.06\%$, and outperforming most existing baselines. Overall, FactorDrive achieves a strong balance between trajectory accuracy and planning safety.

\noindent{\textbf{Results on NAVSIM.}} As shown in Tab.~\ref{tab:navsim_results}, FactorDrive achieves the highest PDMS of 91.0, outperforming the previous best, $AutoDrive\text{-}P^3$, by 0.4 points. It also obtains the best NC and DAC scores of 99.3 and 98.2, respectively, while maintaining a comfort score of 100.0, indicating strong collision avoidance, drivable-area compliance, and trajectory smoothness. Notably, FactorDrive achieves these results without relying on LiDAR, demonstrating strong overall planning performance on NAVSIM.

\subsection{Ablation Study}
\noindent{\textbf{Effect of Different Training Stages.}} As shown in Tab.~\ref{tab:training_stage_ablation}, driving-domain instruction tuning improves the PDMS from 87.9 to 88.4, and QS-GRPO further raises it to 91.0, improving the Driving IT+SFT model by 2.6 points. Notably, our method achieves substantial improvements using fewer than 5,000 hard samples identified by the SFT model. These results demonstrate the effectiveness of RL post-training on targeted hard samples.

\begin{table}[htbp]
\centering
\small
\setlength{\tabcolsep}{0.3pt}
\setlength{\arrayrulewidth}{0.35pt}
\renewcommand{\arraystretch}{1.08}

\begin{tabular*}{\columnwidth}{
@{\extracolsep{\fill}}
l
@{\hspace{2pt}\vrule width 0.35pt\hspace{3pt}}
ccccc
@{\hspace{4pt}\vrule width 0.35pt\hspace{4pt}}
c
@{}
}
\toprule

\textbf{Method}
&
\textbf{NC}\,$\uparrow$
&
\textbf{DAC}\,$\uparrow$
&
\textbf{EP}\,$\uparrow$
&
\textbf{TTC}\,$\uparrow$
&
\textbf{Comf.}\,$\uparrow$
&
\textbf{PDMS}\,$\uparrow$
\\

\midrule

SFT
& 98.7
& 95.6
& 81.4
& 95.7
& \textbf{100.0}
& 87.9
\\

Driving IT+SFT
& \underline{98.8}
& \underline{96.1}
& \underline{81.9}
& \underline{95.9}
& \textbf{100.0}
& \underline{88.4}
\\

Driving IT+SFT+RL
& \textbf{99.3}
& \textbf{98.2}
& \textbf{83.6}
& \textbf{97.6}
& \textbf{100.0}
& \textbf{91.0}
\\

\bottomrule
\end{tabular*}

\caption{
Ablation study of training stages on NAVSIM. Driving IT denotes driving-domain instruction tuning.
}
\label{tab:training_stage_ablation}
\end{table}

\noindent{\textbf{Effect of PCF-CoT Reasoning.}} As shown in Tab.~\ref{tab:pcf_cot_ablation}, under the same driving-domain instruction-tuning and SFT settings, introducing PCF-CoT reasoning has a marginal effect on trajectory L2 error. The results at $1\,\mathrm{s}$ and $2\,\mathrm{s}$ remain unchanged, while the average L2 error increases slightly from $0.28\,\mathrm{m}$ to $0.29\,\mathrm{m}$. In contrast, PCF-CoT reasoning reduces collision rates across all prediction horizons. In particular, the $3\,\mathrm{s}$ collision rate decreases from $0.24\%$ to $0.20\%$, and the average collision rate drops from $0.12\%$ to $0.10\%$, corresponding to a relative reduction of $16.7\%$. These results demonstrate that PCF-CoT reasoning enables the model to leverage planning-critical factors and spatial-physical evidence for safer and more reliable long-horizon planning.

\begin{table}[htbp]
\centering
\small
\setlength{\tabcolsep}{0.7pt}
\setlength{\arrayrulewidth}{0.35pt}
\renewcommand{\arraystretch}{1.08}

\begin{tabular*}{\columnwidth}{
@{\extracolsep{\fill}}
l|cccc|cccc
@{}
}
\toprule

\multirow{2}{*}{\textbf{Method}}
&
\multicolumn{4}{c|}{\textbf{L2 (m)} $\downarrow$}
&
\multicolumn{4}{c}{\textbf{Collision (\%)} $\downarrow$}
\\

\cmidrule(lr){2-5}
\cmidrule(lr){6-9}

&
1s
&
2s
&
3s
&
Avg.
&
1s
&
2s
&
3s
&
Avg.
\\

\midrule

w/o PCF-CoT
& \textbf{0.12}
& \textbf{0.25}
& \textbf{0.48}
& \textbf{0.28}
& 0.03
& 0.08
& 0.24
& 0.12
\\

w/ PCF-CoT
& \textbf{0.12}
& \textbf{0.25}
& 0.49
& 0.29
& \textbf{0.02}
& \textbf{0.07}
& \textbf{0.20}
& \textbf{0.10}
\\

\bottomrule
\end{tabular*}

\caption{
Ablation study of PCF-CoT reasoning on nuScenes.
}
\label{tab:pcf_cot_ablation}
\end{table}

\noindent{\textbf{Effectiveness of QS-GRPO.}} As shown in Tab.~\ref{tab:qs_grpo_ablation}, QS-GRPO improves the PDMS from 90.7 to 91.0 over standard GRPO, with gains of 0.1, 0.4, and 0.4 points in NC, DAC, and EP, respectively, while maintaining a comfort score of 100.0. These results demonstrate that quality-guided search broadens reasoning-path exploration and prioritizes higher-reward paths for policy optimization, mitigating the limited exploration of direct sampling in standard GRPO and further improving planning performance.

\begin{table}[htbp]
\centering
\small
\setlength{\tabcolsep}{2.0pt}
\setlength{\arrayrulewidth}{0.35pt}
\renewcommand{\arraystretch}{1.08}

\begin{tabular*}{\columnwidth}{
@{\extracolsep{\fill}}
l
@{\hspace{2pt}\vrule width 0.35pt\hspace{3pt}}
ccccc
@{\hspace{4pt}\vrule width 0.35pt\hspace{4pt}}
c
@{}
}
\toprule

\textbf{Method}
&
\textbf{NC}\,$\uparrow$
&
\textbf{DAC}\,$\uparrow$
&
\textbf{EP}\,$\uparrow$
&
\textbf{TTC}\,$\uparrow$
&
\textbf{Comf.}\,$\uparrow$
&
\textbf{PDMS}\,$\uparrow$
\\

\midrule

GRPO
& 99.2
& 97.8
& 83.2
& \textbf{97.7}
& \textbf{100.0}
& 90.7
\\

QS-GRPO
& \textbf{99.3}
& \textbf{98.2}
& \textbf{83.6}
& 97.6
& \textbf{100.0}
& \textbf{91.0}
\\

\bottomrule
\end{tabular*}

\caption{
Ablation study of QS-GRPO on NAVSIM.
}
\label{tab:qs_grpo_ablation}
\end{table}

\section{Conclusion}
In this work, we present FactorDrive, an end-to-end autonomous driving framework for adaptive multi-step reasoning driven by planning-critical factors. Through two-stage supervised fine-tuning, FactorDrive first acquires foundational driving knowledge and then learns to perform adaptive multi-step reasoning guided by planning-critical factors and grounded in spatial-physical evidence. QS-GRPO further enhances trajectory planning by using trajectory-level rewards to guide MCTS toward higher-quality reasoning paths and optimizing the policy through GRPO. Evaluations on nuScenes and NAVSIM demonstrate that FactorDrive achieves leading planning performance in both open-loop and closed-loop-oriented settings. We hope this work will inspire further research on scene-adaptive reasoning and reasoning-path optimization for autonomous driving.

{
    \small
    \bibliographystyle{arxiv}
    \bibliography{arxiv}
}


\clearpage
\setcounter{page}{1}
\maketitlesupplementary
\appendix
\section{Data Construction Details}
FactorDrive constructs two data resources tailored to successive stages of supervised fine-tuning. We first build a large-scale Driving-Domain Instruction Corpus to equip a general-purpose VLM with foundational driving knowledge. We then construct the PCF-CoT Dataset to provide trajectory-planning supervision centered on scene-specific planning-critical factors.

\subsection{Driving-Domain Instruction Corpus}
\label{app:instruction_corpus}
Before trajectory planning-oriented SFT, we establish broad driving-domain knowledge by sampling and curating 1,088,695 question--answer (QA) pairs from seven publicly available autonomous-driving question-answering datasets, using only the training split of each dataset. We use seed 42 for sampling and downsampling across all data sources. As shown in Tab.~\ref{tab:instruction_corpus}, these datasets provide diverse and complementary supervision: DriveLM~\cite{sima2024drivelm} connects perception, prediction, and planning; LingoQA~\cite{marcu2024lingoqa} covers driving-video understanding and behavior explanation; OmniDrive~\cite{wang2025omnidrive} focuses on holistic scene understanding, counterfactual reasoning, and trajectory planning; DriveLMM-o1~\cite{ishaq2025drivelmm-o1} provides explicit step-by-step reasoning supervision; nuScenes-QA~\cite{qian2024nuscenes-qa} emphasizes multi-view 3D scene understanding and relational reasoning; Impromptu-VLA~\cite{chi2026impromptu-vla} covers complex and diverse driving scenarios; and STRIDE-QA~\cite{ishihara2026stride-qa} focuses on spatial and spatiotemporal physical reasoning.

We convert the original annotations from all datasets into a unified multimodal dialogue format, in which the user message contains the images and task instruction, and the assistant message contains the corresponding driving-domain response. Through unified sampling and integration, the model acquires the foundational capabilities before learning the PCF-CoT reasoning paradigm, including perceptual understanding, agent motion prediction, trajectory planning, traffic-rule reasoning, counterfactual reasoning, and spatiotemporal physical relation modeling.

\begin{table}[htbp]
\centering
\small
\setlength{\tabcolsep}{3pt}
\renewcommand{\arraystretch}{1.08}

\begin{tabularx}{\columnwidth}{
@{}
X
>{\centering\arraybackslash}p{0.32\columnwidth}
@{}
}
\toprule
\textbf{Dataset} & \textbf{Sampled QA Pairs} \\
\midrule
Impromptu-VLA~\cite{chi2026impromptu-vla} & 230,787 \\
DriveLM~\cite{sima2024drivelm} & 188,991 \\
DriveLMM-o1~\cite{ishaq2025drivelmm-o1} & 18,507 \\
LingoQA~\cite{marcu2024lingoqa} & 206,914 \\
nuScenes-QA~\cite{qian2024nuscenes-qa} & 37,660 \\
OmniDrive~\cite{wang2025omnidrive} & 187,008 \\
STRIDE-QA~\cite{ishihara2026stride-qa} & 218,828 \\
\midrule
\textbf{Total} & \textbf{1,088,695} \\
\bottomrule
\end{tabularx}

\caption{Composition of the Driving-Domain Instruction Corpus.}
\label{tab:instruction_corpus}
\end{table}

\subsection{PCF-CoT Dataset}
PCF-CoT is designed to integrate spatial-physical evidence relevant to future trajectory generation into planning reasoning and to adapt the composition and depth of reasoning paths to the planning demands of different scenes. Rather than exhaustively enumerating all visible elements, the candidate scope of planning-critical factors (PCFs) comprise the subset of scene and motion factors that substantially constrain future ego planning, including road geometry, lane structure and drivable areas, traffic-control elements, critical interacting agents, and ego-motion history. For each sample, PCF-CoT retains only the candidate subset that materially affects longitudinal progress, speed control, the lateral driving corridor, heading change, interaction order, or safety margin.

\subsubsection{Ego-Centric Scene Representation Construction.}
To support reliable PCF annotation, we first align multimodal observations, ego-motion states, map elements, and traffic constraints within a unified ego-centric spatiotemporal representation, and then extract a high-recall set of planning-relevant candidate elements.

\noindent\textbf{Temporal Samples and Unified Coordinate Representation.}
For both nuScenes~\cite{caesar2020nuscenes} and NAVSIM~\cite{dauner2024navsim}, we anchor each sample at the current time $t=0$ and use four temporally ordered front-view camera frames sampled at $2\,\mathrm{Hz}$:
\begin{equation}
    \mathbf{V}=\left\{I_{-1.5},I_{-1.0},I_{-0.5},I_{0}\right\}.
\end{equation}
The ego-motion state includes the historical ego trajectory, the current velocity $\mathbf{v}_0=(v_x,v_y)$, and the current acceleration $\mathbf{a}_0=(a_x,a_y)$. The supervised future ego-trajectory target spans a $4\,\mathrm{s}$ prediction horizon and comprises eight poses sampled at $0.5\,\mathrm{s}$ intervals.

All motion states and map geometries are represented in the current ego-centric coordinate frame. The origin is the ego rear-axle center at $t=0$, with the $x$-axis pointing forward, the $y$-axis pointing left, and positive heading angles measured counterclockwise. Global positions are translated relative to the current ego position and rotated according to the current ego heading; velocity vectors are rotated by the same rotation without translation, and headings are expressed relative to the current ego heading. Historical and future ego trajectories, agent states, lane centerlines, drivable-area boundaries, and traffic-control elements all follow this coordinate convention to avoid reference-frame ambiguity among visual semantics, spatial relations, and trajectory supervision.

For NAVSIM, high-level navigation is directly obtained from the official route instructions and normalized into three categories: Go Straight, Turn Left, and Turn Right. Because nuScenes does not provide equivalent route-level navigation labels, inspired by AutoVLA~\cite{zhou2026autovla}, we construct a navigation target point by accumulating approximately $20\,\mathrm{m}$ along the expert future trajectory polyline. We then assign the same navigation semantics according to the lateral displacement of this target point relative to the trajectory starting point: a lateral displacement of at least $2\,\mathrm{m}$ is labeled Turn Left, a displacement of at most $-2\,\mathrm{m}$ is labeled Turn Right, and all other cases are labeled Go Straight.

\noindent{\textbf{Planning-Relevant Candidate Factor Extraction.}} Raw driving scenes typically contain many agents and map elements that are irrelevant to the current planning decision. To construct the planning-relevant agents, map information, and traffic constraints, we first build a high-recall candidate set using visibility, spatial proximity, and temporal interaction cues. Seed-2.0-Pro~\cite{bytedanceseed2026seed20} subsequently identifies the PCFs that genuinely affect ego planning during structured PCF annotation.

For traffic agents, we consider only objects that are visible in the current front-view camera and belong to planning-related categories, including vehicles, pedestrians, cyclists, traffic cones, and obstacles. An object is retained as a candidate agent if it satisfies at least one of the following conditions: (1) its current 3D bounding box is close to the expert future trajectory corridor; (2) its future trajectory approaches the ego future trajectory; (3) its temporally aligned future positions indicate potential proximity to the ego vehicle; or (4) it lies within the core decision region immediately ahead of the ego vehicle. Candidate agents are ranked by temporally aligned distance, agent-to-ego trajectory distance, and current distance, prioritizing agents that may produce following, crossing, merging, obstacle avoidance, or occlusion effects.

For map information, we retrieve lanes and lane connectors near the current ego vehicle, transform their centerlines into the ego-centric coordinate frame, and retain lanes that are visible in the front-view camera, close to the expert future trajectory, or contain the current ego position. The drivable area is determined by the map polygon containing the current ego vehicle; if no polygon contains it, we select the nearest drivable-area polygon. We then calculate the distances from the ego vehicle to the drivable-area boundaries in the forward, left, and right directions and convert them into front-bumper clearance and left/right body-edge clearance to characterize the geometric feasibility of narrow roads, turning maneuvers, and lateral adjustments.

Traffic constraints include crosswalks, stop lines, road dividers, lane dividers, and traffic lights. We retain only elements that are visible in the front-view camera and intersect the ego future swept corridor, lie close to its boundary, or occur near the trajectory endpoint. For traffic lights, we further associate each light with the relevant stop line or crosswalk according to its relative progress along the future path, reducing the risk of incorrectly treating signals on lateral roads as ego-lane control signals. The main screening rules are summarized in Table~\ref{tab:candidate_screening}. These thresholds are used to construct a high-recall candidate set and do not imply that every retained element is a final PCF; structured PCF annotation further filters candidates according to their actual effects on planning variables.

\begin{table*}[t]
\centering
\small
\setlength{\tabcolsep}{3.5pt}
\setlength{\arrayrulewidth}{0.4pt}
\renewcommand{\arraystretch}{1.08}

\begin{tabularx}{\textwidth}{>{\raggedright\arraybackslash}p{0.19\textwidth}
                              |
                              >{\raggedright\arraybackslash}X}
\toprule
\textbf{Candidate type} & \textbf{Main screening criteria} \\
\midrule
Traffic agents
& We retain front-camera-visible agents from planning-related categories if their current center distance is no larger than $80\,\mathrm{m}$ and their longitudinal position satisfies $x\geq-5\,\mathrm{m}$. The retained agents are further required to either have a current box footprint close to the future ego corridor, have a predicted trajectory close to the future ego trajectory, or lie in the near-front decision zone $0\leq x\leq35\,\mathrm{m}$ and $|y|\leq8\,\mathrm{m}$. \\
\midrule
Lanes and lane connectors
& Lane and lane-connector records are first queried within a $45\,\mathrm{m}$ ego-centered neighborhood. For each retrieved lane, we transform its centerline into the ego frame and keep only front-camera-visible segments that fall within the planning range $-5\leq x\leq80\,\mathrm{m}$ and $|y|\leq20\,\mathrm{m}$, while remaining close to the future ego trajectory. The current ego lane is preserved as a fallback. \\
\midrule
Traffic-control and divider elements
& Traffic-control and divider elements are first queried within a $60\,\mathrm{m}$ ego-centered neighborhood and transformed into the ego frame. We keep elements in the forward planning range $-3\leq x\leq90\,\mathrm{m}$ and $|y|\leq35\,\mathrm{m}$ if they intersect, bound, or lie close to the future ego corridor. This includes stop lines, pedestrian crossings, lane/road dividers, and visible traffic lights whose projected positions fall along the future ego path. \\
\midrule
Drivable areas
& We construct a local drivable region from queried map polygons, prioritizing the region containing the current ego vehicle when such a region is explicitly available; otherwise, the nearest or unioned local drivable region is used. We then compute forward, left, and right boundary distances and derive the corresponding front-bumper and left/right body-edge clearances. \\
\bottomrule
\end{tabularx}

\caption{Main screening rules for extracting planning-relevant candidate factors. All longitudinal and lateral coordinates are measured in the current ego-centric coordinate frame, where $x$ points forward and $y$ points to the left.}
\label{tab:candidate_screening}
\end{table*}

\noindent{\textbf{Teacher-Side Scene Representation.}} To help the annotation model establish correspondence between visual semantics and spatial-physical quantities, we construct a teacher-side scene representation used only for offline data generation.

During offline annotation, we project the 3D bounding boxes of candidate agents, relevant lane centerlines, traffic-control element locations, and the expert future trajectory onto the four historical images. The projected agent boxes establish cross-frame agent correspondence, while the lane-centerline and traffic-control overlays help the annotation model relate visual observations to the corresponding map information. Drivable areas and road boundaries are provided separately to the annotation model as structured ego-centric map information, including boundary distances and body-edge clearances, to support the understanding of road topology and geometric constraints. The projected expert future trajectory delineates the planning-relevant corridor and provides an offline reference for determining the corresponding meta-action and future motion trend. Together, these visual overlays and structured quantities form the teacher-side ego-centric scene representation used for PCF annotation.

At this stage, the expert future trajectory serves as an offline teacher-side reference that helps filter planning-relevant candidates, delimit the planning-relevant reasoning scope, and align the scene representation with the corresponding meta-action and future motion trend. Together with the visual overlays and structured ego-centric quantities, it forms a teacher-side scene representation that provides spatial, motion, and traffic-constraint evidence for subsequent offline PCF annotation.

Importantly, candidate-agent annotations, detailed map geometry, traffic-control annotations, and the projected expert future trajectory are used during the offline construction of PCF-CoT supervision rather than being provided as direct inputs to FactorDrive. During both training and inference, FactorDrive receives only the multimodal input $z$, including the coarse route-level navigation command constructed as described above.

\subsubsection{Planning-Critical Factor Annotation.}

Given the above teacher-side scene representation, we use Seed-2.0-Pro to generate structured PCF annotation, converting raw visual, map, and motion information into the planning-relevant spatial-physical evidence. Structured PCF annotation contains five complementary types of information:

\begin{enumerate}
    \item \textbf{Global Scene Assessment.} It integrates historical images and ego-motion states to describe the current motion state of the ego vehicle, scene type, primary planning constraints, and overall planning demand of the current maneuver.

    \item \textbf{Spatial Constraints.} It covers road curvature, the turning or merging stage, lane topology, drivable-area boundaries, traffic-control constraints, and static corridor restrictions, together with their effects on the feasible region of the future trajectory.

    \item \textbf{Interactions with Critical Agents.} It describes the ego-centric positions, velocities, motion trends, and relationships to the ego future corridor of critical interacting agents, as well as their effects on following distance, yielding, obstacle avoidance, lateral clearance, or collision risk.

    \item \textbf{Counterfactual Action Candidates.} Counterfactual action candidates are constructed only when multiple reasonable actions remain after filtering by traffic rules, route consistency, physical feasibility, and safety constraints. This field records the candidate actions and their primary trade-offs without introducing unnecessary alternatives in scenes where the decision is already largely determined.

    \item \textbf{Ego-Motion and Trajectory Intent.} It describes the final high-level meta-action and the corresponding longitudinal progress, speed trend, lateral displacement, and heading change, providing a consistent planning intent for subsequent Motion Reasoning and future trajectory generation.
\end{enumerate}

For evidence grounding, historical front-view images are used primarily to identify road geometry, traffic-light states, traffic signs, lane markings, and agent semantics, whereas the structured format provides quantitative evidence, including coordinates, distances, velocities, agent motion, and drivable-area clearances. The navigation command is used only as a weak route prior and is not treated as the primary basis for a decision when it conflicts with visual evidence and motion intent. Through this process, raw visual, map, and motion information is organized into compact, verifiable structured PCF annotation that is directly relevant to future trajectory generation.

\subsubsection{PCF-CoT Data Construction.}

Conditioned on the annotated PCFs, we further use Seed-2.0-Pro to adaptively select and organize reasoning units according to the active PCFs in the current scene. PCF-CoT contains the following five reasoning units:

\begin{enumerate}
    \item \textbf{Global Scene Assessment and Reasoning Plan.} This is the initial unit of every CoT. It integrates Global Scene Assessment with the other structured PCFs, identifies the active PCFs that dominate the current planning decision, and determines the reasoning units required in the subsequent path. The Reasoning Plan summarizes only the current ego state, primary planning constraints, and reasoning requirements without prematurely repeating the detailed analysis performed by later units.

    \item \textbf{Scene Spatial Constraint Reasoning.} This unit is activated when road geometry, lane topology, turning or merging structures, drivable-area boundaries, traffic-control constraints, static obstacles, construction areas, or visual occlusions affect ego speed, lateral position, heading, or the driving corridor. Spatial Reasoning analyzes how spatial structure constrains the feasible region of the trajectory. For example, on curved roads or at intersections, it determines the road curvature direction and turning stage and examines their effects on lateral displacement and heading change; in narrow passages, it uses body-edge clearances to assess the feasibility of lateral adjustment.

    \item \textbf{Key Object Prediction and Interaction Reasoning.} This unit is activated when critical interacting agents affect following distance, braking, yielding, lateral clearance, collision risk, heading selection, or visibility. Interaction Reasoning analyzes only the small number of critical agents that directly affect the current planning decision. Using their ego-centric position, relative velocity, motion trend, and relationship to the ego planning corridor, it determines whether each agent acts as a leading vehicle, crossing agent, merging agent, static obstruction, vulnerable road user, or occlusion source.

    \item \textbf{Counterfactual Reasoning.} This unit is activated only when at least two reasonable candidate actions remain after filtering by traffic rules, route consistency, physical feasibility, and safety constraints. Counterfactual Reasoning compares the effects of different actions on safety, driving efficiency, comfort, and trajectory shape and explains the basis for the final action selection. Typical cases include proceeding versus yielding, waiting versus cautious creeping, lane keeping versus lane changing, following versus bypassing, and choosing among different feasible route branches. This reasoning unit is not activated for clear straight driving, ordinary following, or mandatory stopping, where the decision is already largely determined, thereby avoiding superfluous reasoning.

    \item \textbf{Ego Motion and Trajectory Generation Reasoning.} This is the concluding unit of every CoT. It integrates the preceding PCFs and their constraints into a consistent future motion plan, including longitudinal progress, speed change, lateral movement, and heading trend, together with the necessary route consistency, agent clearance, and trajectory smoothness checks. A single overall motion trend is used for scenes with a uniform motion pattern. The future motion is divided into two or three compact stages only when the decision or trajectory pattern actually changes, such as braking followed by cautious creeping, stabilizing after entering a turn, or returning after bypassing an obstacle.
\end{enumerate}

Each reasoning path begins with Global Scene Assessment and Reasoning Plan and concludes with Ego Motion and Trajectory Generation Reasoning. The three intermediate reasoning units are activated as needed. Clear straight-driving scenarios without interaction constraints typically contain only Global Scene Assessment and Reasoning Plan and Ego Motion and Trajectory Generation Reasoning; curved roads, intersections, or traffic-control scenarios additionally activate Scene Spatial Constraint Reasoning; following, crossing, or merging scenarios activate Key Object Prediction and Interaction Reasoning; and Counterfactual Reasoning is activated only when genuine decision trade-offs exist. PCF-CoT therefore adapts reasoning-path composition and depth according to scene-specific PCFs to match the current planning demand, avoiding insufficient analysis in complex scenes and superfluous reasoning in simple scenes.

For output serialization, each reasoning unit is enclosed in an independent \texttt{<think\_step>} tag. The high-level meta-action is enclosed in a \texttt{<meta\_action>} tag and summarizes the high-level driving decision in one concise natural-language sentence. The eight future ego poses are written to the \texttt{<answer>} tag in temporal order from $0.5\,\mathrm{s}$ to $4.0\,\mathrm{s}$. This unified output protocol enables the model to jointly learn PCF selection, reasoning-path organization, meta-action generation, and future trajectory generation within the same autoregressive sequence.

\subsection{Data Quality Assessment}
PCF-CoT adopts a closed-loop quality-control process comprising generation, rule-based format validation, independent VLM-as-Judge evaluation, and iterative revision. Deterministic rules first verify that an output contains only complete and nonempty \texttt{<think\_step>} blocks and remove structurally invalid responses before invoking the judge model.

After rule-based format validation, we use Qwen3.6-Plus~\cite{qwenteam2026qwen36plus}, which differs from the generation model, as an independent VLM-as-Judge to reduce self-evaluation bias~\cite{panickssery2024llm}. The judge model evaluates every sample along the three dimensions:

\begin{enumerate}
    \item \textbf{Planning relevance:} whether the reasoning retains only the evidence that affects the current planning decision;

    \item \textbf{Spatial-physical plausibility:} whether spatial relations, units, coordinate frame, and motion trends remain physically consistent;

    \item \textbf{Output consistency:} whether the meta-action and trajectory trends are consistent with the reasoning outputs.
\end{enumerate}

The VLM-as-Judge produces a structured JSON verdict, rejection reasons, and revision suggestions. Samples that fail the evaluation are returned to Seed-2.0-Pro with the judge feedback for targeted revision, for at most three rounds. Samples that remain invalid are discarded.

\noindent{\textbf{Human Quality Check.}} We assess annotation quality based on planning relevance, spatial-physical plausibility, and output consistency. A sample receives a binary score of 1 only if it satisfies all three criteria and 0 otherwise. Three reviewers independently evaluate 120 sampled PCF-CoT annotations, including 60 from each dataset subset, and the final label is determined by majority vote. The evaluation yields an overall human-verified pass rate of 90.83\%, providing an independent estimate of the reliability of our annotation pipeline.

The resulting PCF-CoT dataset contains 14,802 nuScenes-PCF samples covering 700 nuScenes training scenes and 102,623 NAVSIM-PCF samples covering 1,192 NAVSIM training scenes. To further characterize the adaptive reasoning characteristics of PCF-CoT, we report the reasoning-path statistics for both dataset subsets. As shown in Table~\ref{tab:reasoning_step_distribution}, reasoning paths span two to five steps rather than following a single predefined depth, with three-step and four-step paths accounting for 96.0\% of NAVSIM-PCF samples and 84.5\% of nuScenes-PCF samples. Table~\ref{tab:reasoning_unit_frequency} further shows that Reasoning Plan and Motion Reasoning occur in every sample as the initial and concluding units, whereas the remaining units are activated selectively. These statistics reflect the fine-grained adaptation encoded by PCF-CoT: reasoning-path composition and depth vary with scene-specific planning demands instead of being restricted to a fixed structure or two preset modes.

\begin{table}[htbp]
\centering
\small
\setlength{\tabcolsep}{0.3pt}
\setlength{\arrayrulewidth}{0.35pt}
\renewcommand{\arraystretch}{1.08}

\begin{tabular*}{\columnwidth}{
@{\extracolsep{\fill}}
l
@{\hspace{2pt}\vrule width 0.35pt\hspace{3pt}}
cccc
@{}
}
\toprule
\textbf{Dataset}
& \textbf{2 Steps}
& \textbf{3 Steps}
& \textbf{4 Steps}
& \textbf{5 Steps}
\\
\midrule
NAVSIM
& 3,692
& 69,854
& 28,659
& 418
\\
nuScenes
& 2,200
& 8,971
& 3,530
& 101
\\
\bottomrule
\end{tabular*}

\caption{Distribution of reasoning-path lengths in PCF-CoT.}
\label{tab:reasoning_step_distribution}
\end{table}

\begin{table*}[t]
\centering
\small
\setlength{\tabcolsep}{3.5pt}
\setlength{\arrayrulewidth}{0.4pt}
\renewcommand{\arraystretch}{1.05}

\begin{tabularx}{\textwidth}{
l|
*{5}{c}
}
\toprule
\textbf{Dataset}
& \textbf{Reasoning Plan}
& \textbf{Spatial Reasoning}
& \textbf{Interaction Reasoning}
& \textbf{Counterfactual Reasoning}
& \textbf{Motion Reasoning}
\\
\midrule
NAVSIM
& 102,623 (100.0\%)
& 78,616 (76.6\%)
& 49,156 (47.9\%)
& 654 (0.6\%)
& 102,623 (100.0\%)
\\
nuScenes
& 14,802 (100.0\%)
& 11,238 (75.9\%)
& 4,938 (33.4\%)
& 158 (1.1\%)
& 14,802 (100.0\%)
\\
\bottomrule
\end{tabularx}

\caption{Occurrence counts and proportions of reasoning units in PCF-CoT.}
\label{tab:reasoning_unit_frequency}
\vspace{6pt}
\end{table*}

\section{Two-Stage Supervised Fine-Tuning Details}
\subsection{Driving-Domain Instruction Tuning}

The first stage initializes from Qwen3-VL-8B~\cite{Qwen3-VL} and performs full-parameter instruction tuning on approximately one million driving-domain question--answer (QA) pairs. The primary objective of this stage is not to directly optimize a specific planning benchmark, but to bridge the domain gap between a general-purpose VLM and autonomous driving tasks, thereby equipping the model with the foundational capabilities required for subsequent planning reasoning.

The training corpus covers scene perception, object attribute and relation understanding, traffic-rule reasoning, driving behavior prediction, spatial and temporal relation modeling, counterfactual analysis, and trajectory planning. Because data from different sources have different visual input structures, we do not force all samples into a single fixed input form. Instead, we retain the visual structure required by each original task, including single-frame images, multi-camera views, and multi-frame temporal inputs. All visual content is converted into a unified Qwen3-VL multimodal message format and paired with the corresponding question and target answer to form a user--assistant dialogue.

For data containing explicit image placeholders, we preserve the original interleaved order of images and text. For multi-camera scenes, each camera image is provided as an independent visual block. For temporal data, we preserve the frame order and original sampling frequency.

After serialization with the dialogue template, each training batch contains both the complete multimodal context and the target answer. The loss is applied only to assistant-answer tokens; visual tokens, user questions, and padding positions are excluded from supervision. Unlike task-specific training for a particular output format, this stage preserves the original answer format of each data source, allowing the model to learn a broader range of driving concepts, task formulations, and reasoning patterns and providing a model initialized with driving-domain knowledge for subsequent PCF-CoT fine-tuning.

\subsection{PCF-CoT Supervised Fine-Tuning}

\begin{figure}[t]
    \centering
    \includegraphics[width=0.47\textwidth]{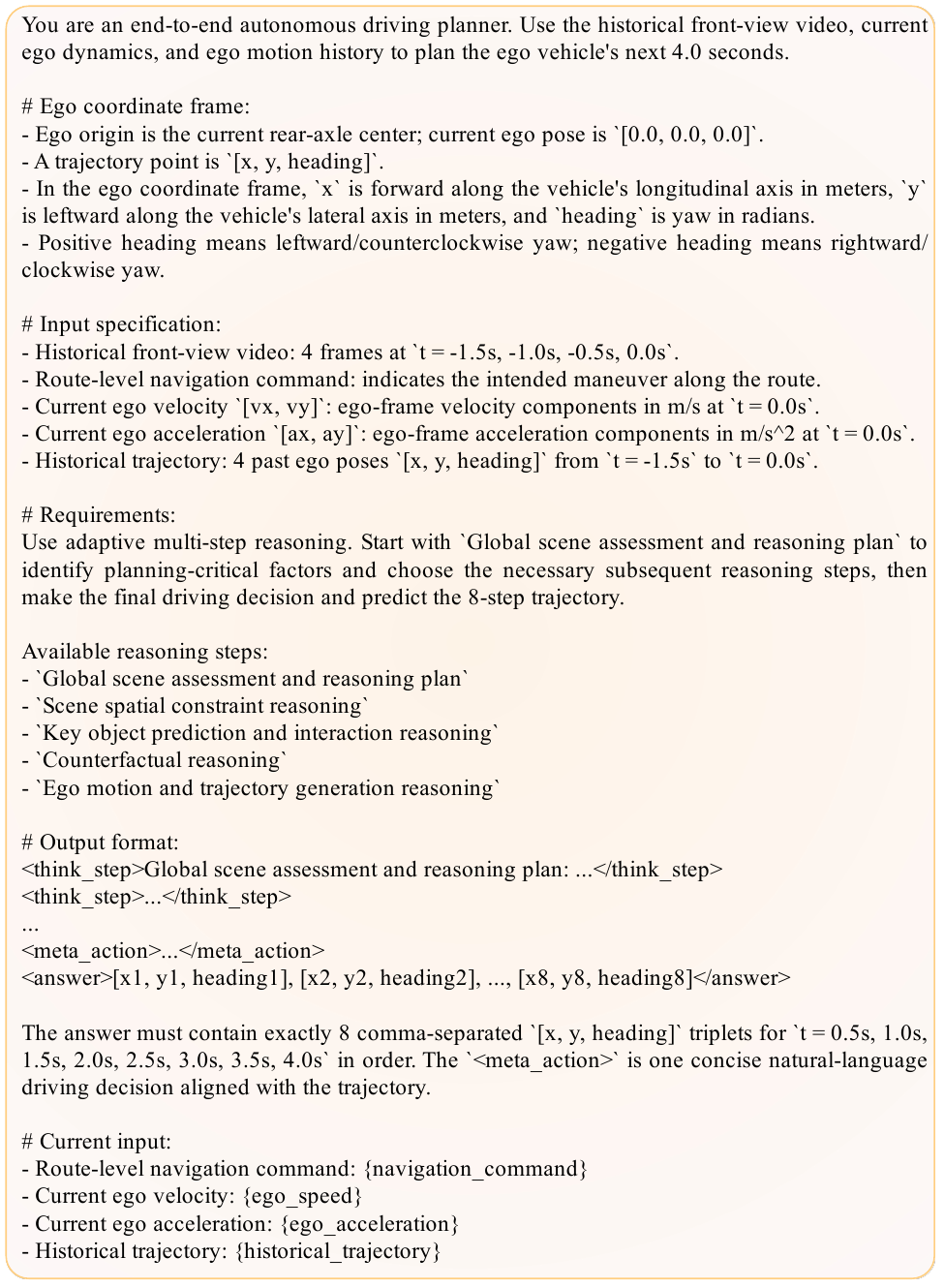}
    \caption{Unified user prompt template used for PCF-CoT supervised fine-tuning.}
    \label{fig:app_sft_prompt}
\end{figure}

The second stage performs planning-specific supervised fine-tuning separately on nuScenes-PCF and NAVSIM-PCF, with both models initialized from the Driving-Domain IT checkpoint obtained in the first stage. Training uses a unified user prompt template. As shown in Fig.~\ref{fig:app_sft_prompt}, the prompt explicitly specifies the ego coordinate frame, historical visual sequence, and trajectory output protocol, and provides the model with four time-stamped front-view images, a route-level navigation command, the current ego velocity and acceleration, and the historical ego trajectory. Based on these inputs, the model first uses Global Scene Assessment and Reasoning Plan to identify the planning-critical factors that affect the current driving decision and determine the necessary subsequent reasoning steps. It then selectively performs Scene Spatial Constraint Reasoning, Key Object Prediction and Interaction Reasoning, Counterfactual Reasoning, and Ego Motion and Trajectory Generation Reasoning according to the specific scene. After completing the reasoning, the model further generates a high-level driving decision and a continuous future trajectory.

\section{QS-GRPO Post-Training Details}
After SFT, the model has acquired adaptive multi-step reasoning capabilities. To further improve trajectory planning through explicit reasoning-path optimization, we propose QS-GRPO, a quality-guided post-training method that explores and reinforces reasoning paths yielding better planning outcomes.

\begin{algorithm*}[!t]
\small
\caption{QS-GRPO Post-Training}
\label{alg:qs-grpo}
\begin{algorithmic}[1]

\Require SFT policy $\pi_{\theta}$, fixed reference policy
$\pi_{\mathrm{ref}}$, hard-sample set
$\mathcal{D}_{\mathrm{hard}}$, group size $G$,
MCTS budget $(K,B,D)$, PUCT coefficient
$c_{\mathrm{puct}}$, quality threshold $\delta_{\mathrm{hq}}$,
and GRPO parameters $(\epsilon,\beta,\delta)$

\Ensure Optimized planning policy $\pi_{\theta}$

\State Initialize $\mathcal{B}(\xi)\leftarrow\emptyset$ for all
$\xi\in\mathcal{D}_{\mathrm{hard}}$

\For{each policy optimization iteration}
    \State Set $\pi_{\theta_{\mathrm{old}}}\leftarrow\pi_{\theta}$,
    sample a mini-batch
    $\mathcal{D}_{b}\subset\mathcal{D}_{\mathrm{hard}}$,
    and initialize $\mathcal{J}_{b}\leftarrow 0$

    \For{each sample $\xi\in\mathcal{D}_{b}$}

        \If{$\mathcal{B}(\xi)\neq\emptyset$}
            \State \textbf{Cached-response reuse:}
            set $\tau_{\mathrm{cache}}\leftarrow\mathcal{B}(\xi)$

            \State Generate and evaluate direct root rollouts
            $\Omega_{\mathrm{roll}}(\xi)$ from
            $\pi_{\theta_{\mathrm{old}}}$

            \State If necessary, perform bounded direct rollout
            fallback to obtain enough unique responses

            \State Set
            $\Omega(\xi)\leftarrow
            \{\tau_{\mathrm{cache}}\}\cup
            \Omega_{\mathrm{roll}}(\xi)$
            and deduplicate complete responses

            \State Select
            $\tau^{*}\leftarrow
            \operatorname*{arg\,max}_{\tau\in\Omega(\xi)}
            r(\tau,\xi)$

            \State Construct
            $\mathcal{G}(\xi)\leftarrow
            \{\tau^{*}\}\cup
            \operatorname{TopConf}_{G-1}
            \bigl(
            \Omega_{\mathrm{roll}}(\xi)
            \setminus\{\tau^{*}\}
            \bigr)$

            \If{$\tau^{*}$ is valid and
                $r(\tau^{*},\xi)>
                r(\tau_{\mathrm{cache}},\xi)$}
                \State Update
                $\mathcal{B}(\xi)\leftarrow\tau^{*}$
            \EndIf

        \Else
            \State \textbf{Quality-guided MCTS:}
            initialize a reasoning tree $\mathcal{T}$ with root $s_{0}$
            and set
            $\Omega_{\mathrm{mcts}}(\xi)\leftarrow\emptyset$,
            $\Omega_{\mathrm{root}}(\xi)\leftarrow\emptyset$

            \For{$k=1,\ldots,K$}
                \State \textbf{Selection:}
                recursively maximize the PUCT score from $s_{0}$
                until reaching a non-terminal leaf $s$ or depth $D$

                \State \textbf{Expansion and rollout:}
                sample at most $B$ complete responses from
                $\pi_{\theta_{\mathrm{old}}}
                (\cdot\mid s,z_{\xi})$

                \For{each generated response $\tau$}
                    \State Evaluate $r(\tau,\xi)$ and add $\tau$ to
                    $\Omega_{\mathrm{mcts}}(\xi)$

                    \If{$s=s_{0}$}
                        \State Add $\tau$ to
                        $\Omega_{\mathrm{root}}(\xi)$
                    \EndIf

                    \State Expand $\mathcal{T}$ with the first newly
                    generated complete reasoning step, if one exists

                    \State Backpropagate $r(\tau,\xi)$ along the
                    visited path and update $N$, $W$, and $Q$
                \EndFor

                \If{$|\Omega_{\mathrm{mcts}}(\xi)|>G$ and
                    $\max_{\tau\in\Omega_{\mathrm{mcts}}(\xi)}
                    r(\tau,\xi)\geq\delta_{\mathrm{hq}}$}
                    \State \textbf{break}
                \EndIf
            \EndFor

            \State Deduplicate the complete responses in
            $\Omega_{\mathrm{mcts}}(\xi)$ and
            $\Omega_{\mathrm{root}}(\xi)$

            \State Select
            $\tau^{*}\leftarrow
            \operatorname*{arg\,max}_{\tau\in
            \Omega_{\mathrm{mcts}}(\xi)}
            r(\tau,\xi)$

            \State Construct
            $\mathcal{G}(\xi)\leftarrow
            \{\tau^{*}\}\cup
            \operatorname{TopConf}_{G-1}
            \bigl(
            \Omega_{\mathrm{root}}(\xi)
            \setminus\{\tau^{*}\}
            \bigr)$

            \If{$|\mathcal{G}(\xi)|<G$}
                \State Perform bounded direct root rollouts and append
                the highest-confidence non-duplicate responses until
                $|\mathcal{G}(\xi)|=G$
            \EndIf

            \If{$\tau^{*}$ is valid and
                $r(\tau^{*},\xi)\geq\delta_{\mathrm{hq}}$}
                \State Update
                $\mathcal{B}(\xi)\leftarrow\tau^{*}$
            \EndIf
        \EndIf

        \State Recompute old-policy token log-probabilities for
        $\mathcal{G}(\xi)$, calculate group-relative advantages,
        and accumulate the clipped QS-GRPO objective with KL
        regularization into $\mathcal{J}_{b}$
    \EndFor

    \State Update $\theta$ by maximizing $\mathcal{J}_{b}$
\EndFor

\State \Return $\pi_{\theta}$

\end{algorithmic}
\end{algorithm*}

The complete QS-GRPO pipeline consists of Hard-Sample Selection, quality-guided MCTS, a High-Quality Response Buffer, Confidence-Aware Group Construction, and the final group-relative policy optimization. Algorithm~\ref{alg:qs-grpo} summarizes the overall optimization flow, while the following subsections provide the precise execution details for within-iteration buffer updates, candidate-pool deduplication, and confidence-aware group construction.

\subsection{Hard-Sample Selection}
We first use the corresponding PCF-CoT SFT model to perform planning evaluation on the full training split and construct the hard-sample subset $\mathcal{D}_{\mathrm{hard}}$ according to its planning performance. QS-GRPO performs post-training only on $\mathcal{D}_{\mathrm{hard}}$, concentrating the search budget on scenarios that the initial PCF-CoT SFT policy does not yet solve reliably and that offer greater optimization potential.

For any hard sample $\xi$, MCTS constructs a reasoning-state tree conditioned on its multimodal driving input $z_\xi$. A node $s$ represents the reasoning prefix already generated by the model, which consists of zero or more complete \texttt{<think\_step>} reasoning units. A tree-edge action $u$ represents the next complete reasoning unit appended after the current prefix. Upon reaching a leaf node, the old policy continues generating the remaining reasoning, meta-action, and future trajectory to obtain a complete response $\tau$, where the meta-action and future trajectory form the terminal output.

\subsection{Quality-Guided MCTS for Reasoning Path Search}
For each candidate edge $(s,u)$ under state $s$, we maintain three statistics:
\begin{itemize}
    \item $N(s,u)$: the visit count of the reasoning branch;
    \item $W(s,u)$: the cumulative complete-response reward obtained through the branch;
    \item $Q(s,u)$: the average total reward $r(\tau,\xi)$ obtained through the branch.
\end{itemize}

The total visit count of the parent node is defined as
\begin{equation}
N(s)=\sum_{u\in\mathcal U(s)}N(s,u),
\label{eq:app_parent_visits}
\end{equation}
where $\mathcal U(s)$ denotes the set of candidate actions already expanded under state $s$. For an unvisited edge, we initialize $N(s,u)=0$, $W(s,u)=0$, and $Q(s,u)=0$.

\subsubsection{PUCT Selection.} During selection, the search starts from the root node and uses the PUCT criterion~\cite{silver2017mcts} at each intermediate node to select the next reasoning branch:
\begin{equation}
\resizebox{0.95\columnwidth}{!}{$\displaystyle
u^{*}=\underset{u\in\mathcal U(s)}{\arg\max}\biggl[
Q(s,u)+c_{\mathrm{puct}}P_{\mathrm{old}}(u\mid s,z_\xi)
\frac{\sqrt{\max(N(s),1)}}{1+N(s,u)}
\biggr].$}
\label{eq:app_puct}
\end{equation}
where $P_{\mathrm{old}}(u\mid s,z_\xi)$ denotes the relative generation probability assigned by the old policy to candidate reasoning unit $u$. The first term encourages the search to exploit reasoning branches that have obtained higher planning rewards, while the second combines the old-policy probability with the branch visit count to preserve exploration opportunities for reasoning actions that have not been sufficiently evaluated. We set $c_{\mathrm{puct}}=1.5$ in the experiments.

\subsubsection{Expansion, Evaluation, and Backpropagation.}
Each MCTS simulation consists of selection, expansion and rollout, and backpropagation.

\begin{enumerate}
    \item \textbf{Selection.} The search starts from the root node, selects the reasoning action with the highest PUCT score at each intermediate node, and recursively follows the corresponding tree edge until reaching a non-terminal leaf that has not been fully expanded. This process exploits high-value reasoning branches while preserving the necessary exploration opportunities for candidates that have not been sufficiently evaluated.

    \item \textbf{Expansion and Rollout.} Upon reaching a leaf node, the old policy $\pi_{\theta_{\mathrm{old}}}$ generates at most $B=8$ complete candidate suffixes from the reasoning prefix associated with that node. Each candidate suffix may contain one or more subsequent reasoning units, followed by the final meta-action and future trajectory.

    We add the first complete reasoning unit newly generated after the current prefix to the search tree as a tree-edge action. Candidates sharing the same first newly generated reasoning unit are merged into the same tree edge, while their complete responses are retained and scored independently. If a candidate directly generates the terminal output without producing a new reasoning unit, its complete response still participates in candidate-pool construction and reward computation, but no new reasoning edge is expanded.

    The reasoning prefix of the leaf node and each candidate suffix are then combined into a complete response $\tau$, which is evaluated using the total complete-response reward $r(\tau,\xi)$ of the corresponding benchmark. Because the reward is computed from the complete response, the final trajectory quality provides supervision for the entire reasoning path.

    \item \textbf{Backpropagation.} For each evaluated response, its reward is backpropagated without discounting along the corresponding visited path. For every edge $(s,u)$ on the path, the statistics are updated as
    \begin{equation}
    N(s,u)\leftarrow N(s,u)+1,
    \label{eq:app_backpropagation_visit}
    \end{equation}
    \begin{equation}
    W(s,u)\leftarrow W(s,u)+r(\tau,\xi),
    \label{eq:app_backpropagation_reward}
    \end{equation}
    \begin{equation}
    Q(s,u)=\frac{W(s,u)}{N(s,u)},
    \label{eq:app_backpropagation_value}
    \end{equation}
    
\end{enumerate}

\noindent{\textbf{Search Budget and Early Termination.}} For each hard sample that has not yet obtained a high-quality cached response, we execute at most $K=6$ MCTS simulations. Each node expands at most $B=8$ candidate actions, and the maximum search depth is $D=5$ reasoning units.

During search, we separately record the candidate set generated directly from the root node, denoted by $\Omega_{\mathrm{root}}(\xi)$. The search terminates early when the collected MCTS responses satisfy the candidate-count condition in Algorithm~\ref{alg:qs-grpo} and the highest candidate reward reaches the benchmark-specific high-quality reward threshold $\delta_{\mathrm{hq}}$. Otherwise, simulations continue until the search budget is exhausted. This MCTS budget is separate from any additional direct root-node sampling used after search to obtain enough unique responses for group construction.

\noindent{\textbf{High-Quality Response Buffer.}} To reduce repeated full-tree searches for samples that have already obtained a reliable solution, we maintain a High-Quality Response Buffer $\mathcal B(\xi)$ for each hard sample. Each sample stores at most one response, and a response is stored only when its total reward is no lower than $\delta_{\mathrm{hq}}$ for the corresponding benchmark.

The candidate-generation branch used in the current iteration is determined by the buffer state at the beginning of that iteration. For a sample without a cached response at iteration start, we perform a full MCTS search and select the highest-reward response $\tau_{\mathrm{mcts}}^{*}$ from $\Omega_{\mathrm{mcts}}(\xi)$. If its reward reaches $\delta_{\mathrm{hq}}$, it is written to $\mathcal B(\xi)$ for reuse in subsequent iterations; otherwise, the buffer remains empty. Regardless of whether the response is cached, the current iteration continues to use the no-cache candidate-generation branch.

For a sample with an existing cached response at iteration start, no additional full-tree search is performed. Instead, the historical cached response is compared with newly generated direct rollouts $\Omega_{\mathrm{roll}}(\xi)$. If no rollout has a strictly higher reward, the historical response remains in the buffer. If a rollout response strictly outperforms it, the rollout response immediately replaces the historical cache. The superseded historical response is removed from the effective candidate pool of the current iteration. Since the replacing response already belongs to $\Omega_{\mathrm{roll}}(\xi)$, it remains available for current optimization without being inserted a second time. The resulting buffer state is carried forward to subsequent iterations.

Thus, branch selection, within-iteration candidate-pool update, and buffer persistence have distinct timing. Branch selection depends only on the buffer state at iteration start, a buffer replacement takes effect immediately in the current effective candidate pool, and the updated buffer is then reused in later iterations.

\noindent{\textbf{Confidence-Aware Group Construction.}} Algorithm~\ref{alg:qs-grpo} provides a compact summary of group construction, while the following description specifies the precise candidate-pool semantics after within-iteration buffer updates and exact-response deduplication. Candidate-pool construction follows Eq.~(3) of the main paper. Its branch condition refers to the buffer state at the beginning of the current training iteration. For a sample without a cache at iteration start, the candidate pool contains the highest-reward MCTS response $\tau_{\mathrm{mcts}}^{*}$ and direct rollouts generated from the root node. For a sample with a cache at iteration start, the historical cache is first compared with the newly generated direct rollouts. If the cache is retained, the effective candidate pool contains the retained response and the current rollout responses. If a current rollout replaces the historical cache, the superseded response is removed, and the effective candidate pool consists only of the current rollout responses.

Candidate deduplication is performed after the within-iteration buffer comparison and removal of any superseded cached response. Two candidates are treated as duplicates only when their complete serialized responses, including the reasoning path, meta-action, and future trajectory, match exactly. If fewer than $G$ unique eligible responses remain after deduplication, additional responses are generated through direct autoregressive sampling from the current old policy. In the no-cache branch, these responses are sampled from the root state; in the cached branch, they are added to the current direct-rollout set. Sampling continues until enough unique candidates are obtained.

The final training group $\mathcal G(\xi)$ is constructed according to Eq.~(5) of the main paper. Here, the candidate pool refers to the deduplicated effective candidate pool obtained after the within-iteration buffer comparison and removal of any superseded cached response. We first retain its highest-reward response as $\tau^{*}$ and then select the remaining $G-1$ responses according to the length-normalized confidence assigned by the current old policy. All candidates that remain eligible for $\operatorname{TopConf}_{G-1}$ are evaluated under the same $\pi_{\theta_{\mathrm{old}}}$.

This rule has three concrete outcomes. In the no-cache branch, $\tau^{*}$ may be either $\tau_{\mathrm{mcts}}^{*}$ or a higher-reward root-node rollout. If the historical cache is retained, it supplies the reward-selected response, and the remaining responses are selected by confidence from the current rollouts. If a current rollout replaces the historical cache, the superseded response is discarded, and all $G$ training responses originate from the current rollout set: one is selected by reward and the remaining $G-1$ are selected by confidence. We set $G=8$ in the experiments.

This construction exposes both high-quality planning paths and responses that the current policy is likely to generate. High-confidence but low-reward candidates receive lower group-relative advantages and are therefore suppressed during policy optimization; confidence itself is not treated as a positive training target. A retained cached response enters the group through reward-based selection, whereas a superseded cached response is removed and does not participate in confidence-based selection.

\subsection{QS-GRPO Optimization Objective}
QS-GRPO is optimized using group-relative advantages and a clipped policy objective. The numerical-stability constant used in the advantage normalization in Eq.~(6) of the main paper is set to $\delta=10^{-4}$. The reference policy $\pi_{\mathrm{ref}}$ in Eq.~(7) of the main paper is fixed to the PCF-CoT SFT model used to initialize QS-GRPO, and the KL regularization coefficient is set to $\beta=0.001$. In addition, the KL divergence is estimated using a non-negative token-level estimate:

\begin{equation}
d_{i,t} = \log\pi_{\mathrm{ref}}\left(y_{i,t}\mid y_{i,<t},z_\xi\right) - \log\pi_\theta\left(y_{i,t}\mid y_{i,<t},z_\xi\right).
\label{eq:app_token_kl_difference}
\end{equation}

The KL divergence estimate over the training group is
\begin{equation}
\widehat D_{\mathrm{KL}}
=\frac{1}{G}\sum_{i=1}^{G}\frac{1}{T_i}
\sum_{t=1}^{T_i}
\left[\exp(d_{i,t})-d_{i,t}-1\right],
\label{eq:app_token_kl}
\end{equation}
where $G$ is the training group size and $T_i$ is the valid token count of response $\tau_i$. This estimate is always non-negative and becomes zero when the current policy and reference policy are identical.

In the actual optimization, the response-level importance ratio in Eq.~(7) of the main paper is evaluated through token-level importance ratios, and KL regularization is likewise computed at the token level. The group-relative advantage of each response is shared by all its valid tokens. The loss is first averaged over the valid tokens of each response and then averaged over the training group, preventing response length from introducing an additional effect on the gradient scale.

For a retained response reused from the High-Quality Response Buffer, token-level log-probabilities are recomputed under the current old policy $\pi_{\theta_{\mathrm{old}}}$ by teacher forcing. These values provide the old-policy likelihood required by the token-level importance ratio in the current GRPO update. This re-evaluation does not make the historical response an on-policy sample and should not be interpreted as eliminating the mismatch between the policy that originally generated the response and the current old policy. Confidence ranking and old-policy likelihood recomputation serve different purposes: confidence ranks candidates eligible for $\operatorname{TopConf}_{G-1}$, whereas the recomputed token-level likelihood provides the old-policy probability required for policy optimization after the group has been constructed.

\section{Experimental Details}
\subsection{Benchmarks and Metrics}
\noindent{\textbf{nuScenes.}} nuScenes~\cite{caesar2020nuscenes} is a large-scale multimodal autonomous driving dataset for complex urban traffic scenarios. It contains 1,000 driving scenes of approximately 20 seconds each, divided into training, validation, and test splits with 700, 150, and 150 scenes, respectively. Following the ST-P3 evaluation setting~\cite{hu2022stp3}, we conduct open-loop trajectory planning evaluation on the validation split and report L2 errors at the $1\,\mathrm{s}$, $2\,\mathrm{s}$, and $3\,\mathrm{s}$ horizons and their average, together with the collision rate at each horizon. Using four historical frames and six future frames, we construct temporal samples from the 6,019 frames in the validation split. Specifically, we discard the first three frames and the last six frames of each of the 150 validation scenes because they cannot provide complete historical and future context, respectively, yielding 4,669 evaluation frames.

The L2 error computes the Euclidean distance between the predicted trajectory point and the ground-truth ego trajectory point to measure trajectory prediction accuracy. The collision rate measures the proportion of samples in which the predicted ego trajectory spatially conflicts with other traffic participants to evaluate planning safety. Lower values indicate better performance for both metrics.

\noindent{\textbf{NAVSIM.}} NAVSIM~\cite{dauner2024navsim} constructs a non-reactive simulation environment from real-world driving logs and evaluates planning safety, driving efficiency, and comfort by executing the planned trajectory over a fixed $4\,\mathrm{s}$ prediction horizon. We evaluate on the 12,146 samples in the navtest split.

We adopt the PDM Score (PDMS) as the overall planning metric. It consists of no-at-fault collision (NC), drivable area compliance (DAC), ego progress (EP), time-to-collision (TTC), and comfort (C), and is defined as
\begin{equation}
\mathrm{PDMS}=\mathrm{NC}\times\mathrm{DAC}\times\frac{5\,\mathrm{EP}+5\,\mathrm{TTC}+2\,\mathrm{C}}{12}.
\label{eq:app_pdms}
\end{equation}
NC and DAC serve as multiplicative safety constraints that directly penalize collisions and departures from the drivable area. EP, TTC, and C measure effective driving progress, potential collision risk, and trajectory smoothness, respectively. Each component metric and the final PDMS are normalized to $[0,1]$, with higher values indicating better planning performance. Following the reporting convention of existing NAVSIM methods~\cite{liao2025diffusiondrive,zhou2026autovla,ye2026autodrivetp3}, we multiply these metrics by 100 for presentation.

For inference-time evaluation on both benchmarks, we deploy the models using vLLM 0.11.0~\cite{kwon2023vllm} and perform batched inference with the temperature set to 0 and top-$p$ set to 1. We conduct a single complete evaluation run for each benchmark, generating one response per sample to ensure consistent and reproducible results.

\subsection{Implementation Details}
\subsubsection{Hard-Sample Selection.} We first use the corresponding PCF-CoT SFT model to perform planning evaluation on the full training split and construct the hard-sample subset $\mathcal D_{\mathrm{hard}}$ according to its planning performance. For nuScenes, we select 6,407 samples satisfying $\mathrm{L2}_{3\mathrm{s}}>0.5\,\mathrm{m}$; for NAVSIM, we select 4,910 samples satisfying $\mathrm{PDMS}<0.80$. QS-GRPO performs post-training only on $\mathcal D_{\mathrm{hard}}$, concentrating the search budget on scenarios that the current policy has not yet solved reliably and that offer greater optimization potential.

\begin{figure}[h]
\centering
\begin{tabular}{@{}c@{\hspace{2pt}}c@{}}
\includegraphics[width=0.49\columnwidth]{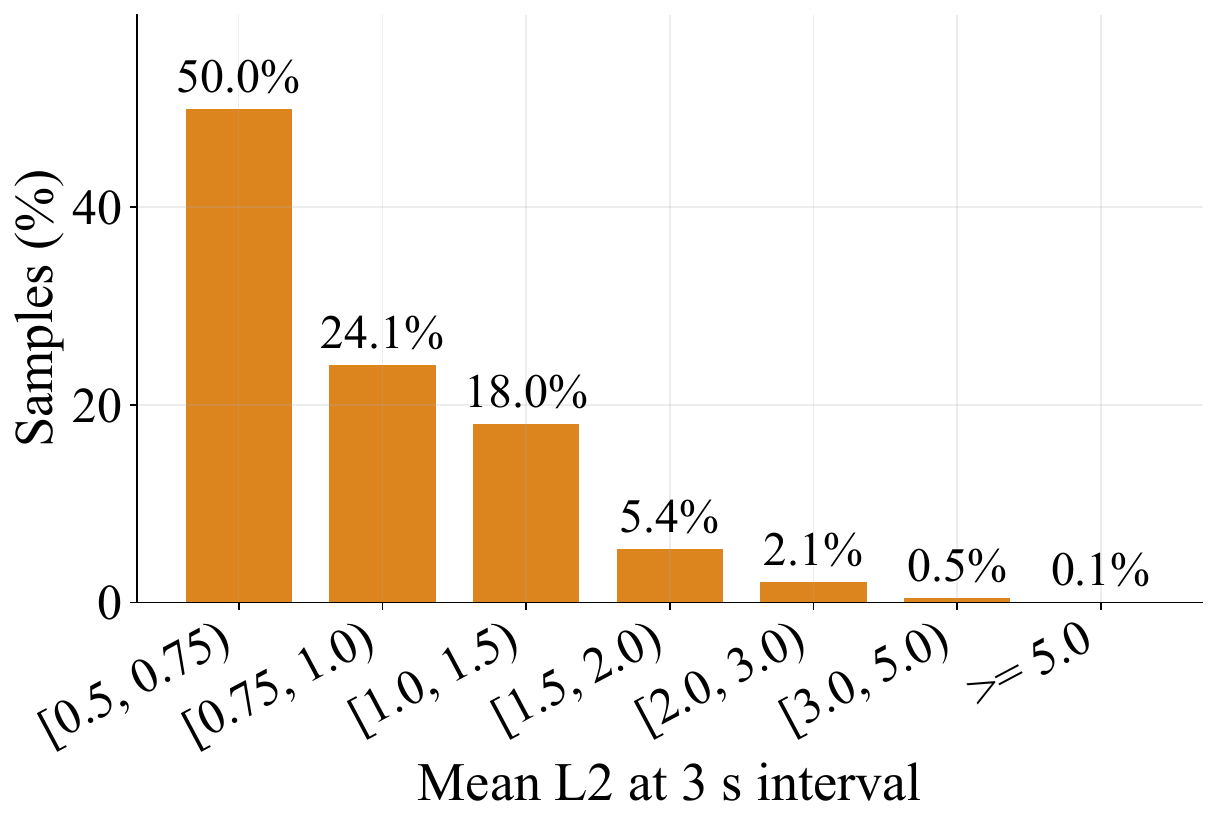} &
\includegraphics[width=0.49\columnwidth]{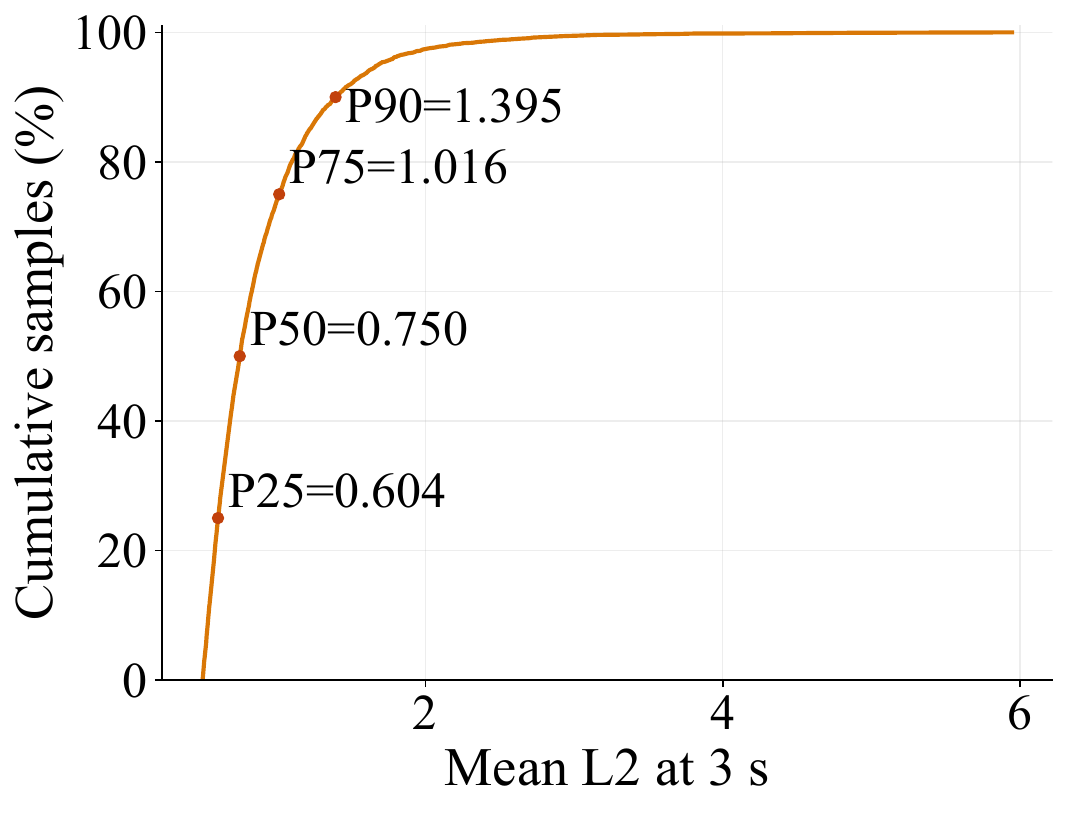} \\
\textnormal{(a)} & \textnormal{(b)}
\end{tabular}
\caption{Distribution of $\mathrm{L2}_{3\mathrm{s}}$ among the selected nuScenes hard samples. (a) Sample proportions across error intervals. (b) Empirical cumulative distribution function with percentile markers.}
\label{fig:app_nuscenes_hard_distribution}
\end{figure}

\begin{figure}[h]
\centering
\begin{tabular}{@{}c@{\hspace{2pt}}c@{}}
\includegraphics[width=0.49\columnwidth]{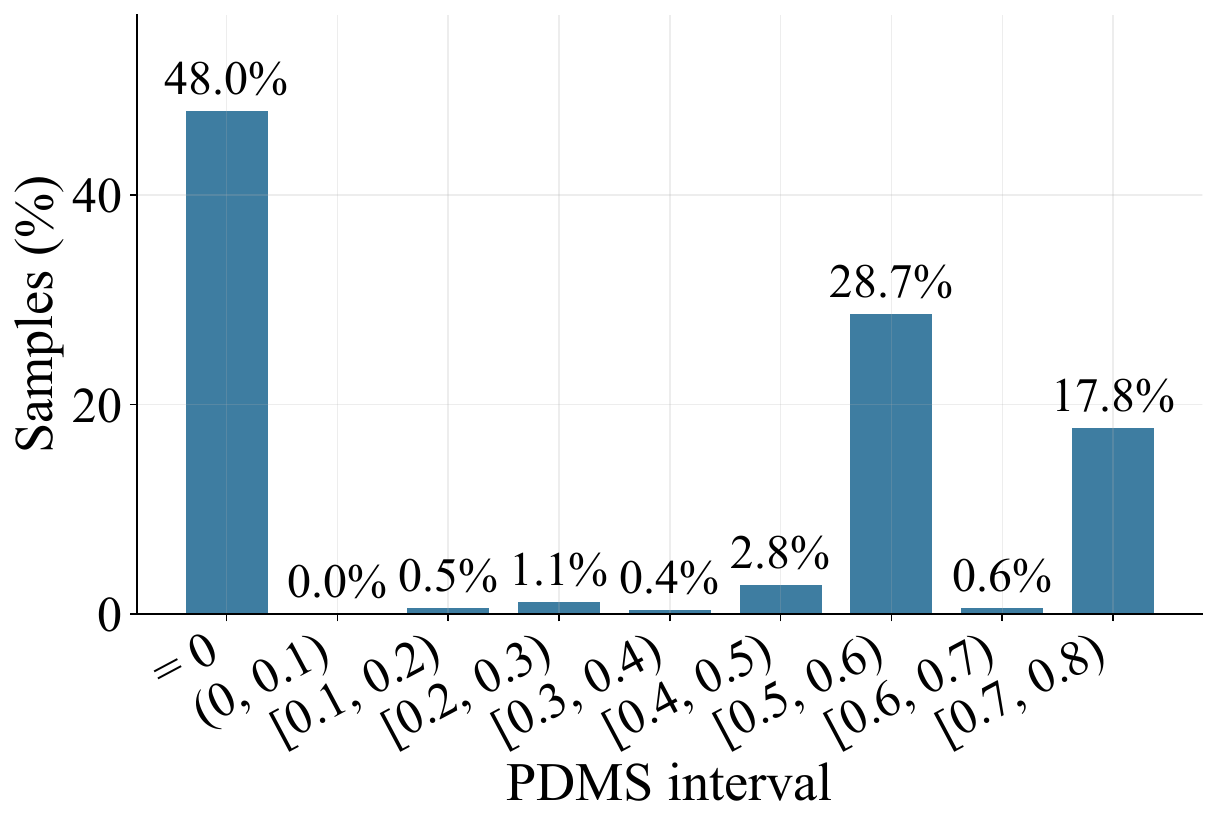} &
\includegraphics[width=0.49\columnwidth]{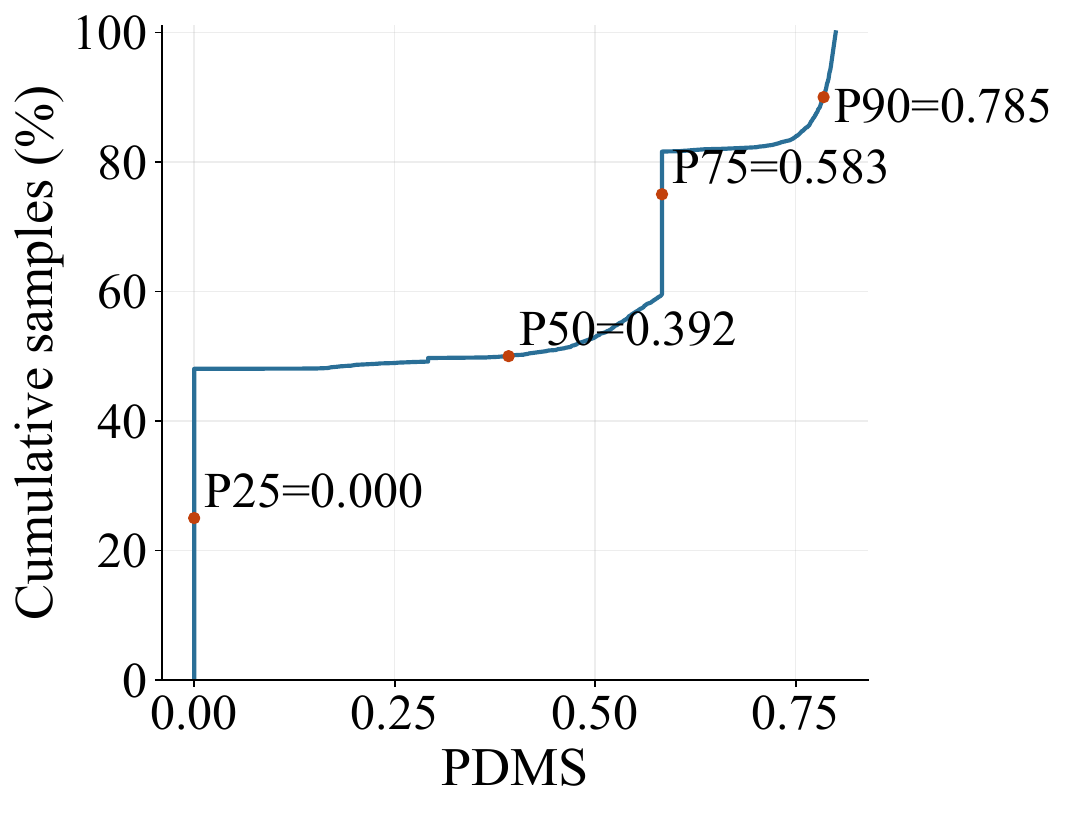} \\
\textnormal{(a)} & \textnormal{(b)}
\end{tabular}
\caption{Distribution of PDMS among the selected NAVSIM hard samples. (a) Sample proportions across score intervals. (b) Empirical cumulative distribution function with percentile markers.}
\label{fig:app_navsim_hard_distribution}
\end{figure}

Figs.~\ref{fig:app_nuscenes_hard_distribution} and~\ref{fig:app_navsim_hard_distribution} characterize the score distributions within the selected hard-sample subsets. For nuScenes, 50.0\% of the selected samples fall in $[0.5,0.75)\,\mathrm{m}$, while the 75th and 90th percentiles reach $1.016\,\mathrm{m}$ and $1.395\,\mathrm{m}$, respectively, indicating that the subset includes both near-threshold cases and a tail of larger long-horizon errors. For NAVSIM, 48.0\% of the selected samples have zero PDMS, while 28.7\% and 17.8\% fall in $[0.5,0.6)$ and $[0.7,0.8)$, respectively, covering both zero-score failures and cases close to the selection boundary. Although L2 and PDMS are not directly comparable, both distributions show that hard-sample selection directs the QS-GRPO search budget toward unresolved planning cases across a range of failure severities rather than only a narrow score interval.

\begin{table*}[t]
\centering
\small
\setlength{\tabcolsep}{12.5pt}
\setlength{\arrayrulewidth}{0.4pt}
\renewcommand{\arraystretch}{1.08}
\begin{tabularx}{\textwidth}{l|c|*{2}{c}|*{2}{c}}
\toprule
\multirow{2}{*}{\textbf{Setting}}
& \multirow{2}{*}{\textbf{\shortstack{Driving-\\Domain IT}}}
& \multicolumn{2}{c|}{\textbf{PCF-CoT SFT}}
& \multicolumn{2}{c}{\textbf{QS-GRPO}}
\\
\cmidrule(lr){3-4}
\cmidrule(lr){5-6}
& & \textbf{nuScenes} & \textbf{NAVSIM} & \textbf{nuScenes} & \textbf{NAVSIM}
\\
\midrule
Epochs & 1 & 3 & 4 & 3 (2,400 steps) & 5 (3,065 steps) \\
Batch size & 128 & 16 & 16 & 8 & 8 \\
Optimizer & AdamW & AdamW & AdamW & AdamW & AdamW \\
Learning rate & $5\times10^{-6}$ & $1\times10^{-5}$ & $1\times10^{-5}$ & $1\times10^{-6}$ & $1\times10^{-6}$ \\
Warmup ratio & 0.03 & 0.03 & 0.03 & 0.03 & 0.03 \\
Maximum gradient norm & 1 & 1 & 1 & 1 & 1 \\
KL coefficient ($\beta$) & -- & -- & -- & 0.001 & 0.001 \\
Rollout group size ($G$) & -- & -- & -- & 8 & 8 \\
Image setting & \shortstack{adaptive} & $[4,3,384,672]$ & $[4,3,384,672]$ & $[4,3,384,672]$ & $[4,3,384,672]$ \\
\bottomrule
\end{tabularx}
\caption{Training configurations for Driving-Domain IT, PCF-CoT SFT, and QS-GRPO.}
\label{tab:app_training_details}
\end{table*}

The benchmark-specific criteria retain 6,407 of 20,438 nuScenes training samples and 4,910 of 103,288 NAVSIM training samples, corresponding to 31.3\% and 4.8\%, respectively. Thus, the nuScenes criterion covers a broader set of samples with large long-horizon trajectory errors, whereas the NAVSIM criterion concentrates post-training on a smaller subset with low overall planning scores. These proportions are not directly comparable as measures of dataset difficulty because L2 and PDMS have different definitions and numerical scales.

\subsubsection{Reward Design.} We define the trajectory-level planning reward of a complete response $\tau$ as
\begin{equation}
r(\tau,\xi)=\lambda_{\mathrm{fmt}}R_{\mathrm{fmt}}+\lambda_{\mathrm{traj}}R_{\mathrm{traj}}.
\label{eq:app_planning_reward}
\end{equation}
We set $\lambda_{\mathrm{fmt}}=0.2$ and $\lambda_{\mathrm{traj}}=1.0$, corresponding to a weight ratio of $1:5$.

\noindent{\textbf{Format Reward.}} The format reward $R_{\mathrm{fmt}}\in\{0,1\}$ ensures that the model response can be parsed deterministically. A valid response must contain, in order, one or more structurally complete \texttt{<think\_step>} units, exactly one non-empty \texttt{<meta\_action>}, and exactly one \texttt{<answer>}. The \texttt{<answer>} must contain exactly eight parseable $(x,y,\psi)$ trajectory points.

\noindent{\textbf{nuScenes Trajectory Reward.}} Let $\hat{\mathbf p}_k$ and $\mathbf p_k^{\mathrm{gt}}$ denote the $k$-th predicted position and ground-truth position, respectively, both represented as two-dimensional coordinates.

For an evaluation horizon containing $H$ trajectory points, the average trajectory error is defined as $\mathrm{L2}_{H}=\frac{1}{H}\sum_{k=1}^{H}\|\hat{\mathbf p}_k-\mathbf p_k^{\mathrm{gt}}\|_2$, and the final displacement error is defined as $\mathrm{FDE}_{H}=\|\hat{\mathbf p}_H-\mathbf p_H^{\mathrm{gt}}\|_2$.
The corresponding average trajectory error reward is defined as
\begin{equation}
R_{\mathrm{L2}}=\exp\!\left(-\frac{\mathrm{L2}_{H}}{\alpha_{\mathrm{L2}}}\right).
\label{eq:app_l2_reward}
\end{equation}
The final displacement error reward is defined as
\begin{equation}
R_{\mathrm{FDE}}=\exp\!\left(-\frac{\mathrm{FDE}_{H}}{\alpha_{\mathrm{FDE}}}\right).
\label{eq:app_fde_reward}
\end{equation}
We set $\alpha_{\mathrm{L2}}=\alpha_{\mathrm{FDE}}=1.0$.

nuScenes follows a $3\,\mathrm{s}$ evaluation horizon and uses the first six points of the eight-point output trajectory, yielding $H=6$. Its trajectory-quality reward is
\begin{equation}
R_{\mathrm{traj}}^{\mathrm{nuScenes}}=R_{\mathrm{L2}}+R_{\mathrm{FDE}}.
\label{eq:app_nuscenes_trajectory_reward}
\end{equation}

\noindent{\textbf{NAVSIM Trajectory Reward.}} NAVSIM uses the complete eight-point trajectory over $4\,\mathrm{s}$, yielding $H=8$. In addition to $R_{\mathrm{L2}}$, we follow AutoVLA~\cite{zhou2026autovla} and introduce a PDMS reward, defined as $R_{\mathrm{PDMS}}=\mathrm{PDMS}$. Both $R_{\mathrm{L2}}$ and $R_{\mathrm{PDMS}}$ lie in $[0,1]$. The NAVSIM trajectory-quality reward is
\begin{equation}
R_{\mathrm{traj}}^{\mathrm{NAVSIM}}=R_{\mathrm{L2}}+R_{\mathrm{PDMS}}.
\label{eq:app_navsim_trajectory_reward}
\end{equation}

For quality-guided MCTS and the High-Quality Response Buffer, we set the benchmark-specific high-quality threshold applied to the total reward $r(\tau,\xi)$ to $\delta_{\mathrm{hq}}=1.8$ for nuScenes and $\delta_{\mathrm{hq}}=2.05$ for NAVSIM.

\noindent{\textbf{Training Details.}} The main training settings for all stages are summarized in Tab.~\ref{tab:app_training_details}. For Driving-Domain IT, adaptive indicates that input images are dynamically resized rather than converted to a fixed tensor shape, with the pixel budget constrained by setting \texttt{max\_pixels} to 262,144. All three training stages use random seed 42.

\section{Additional Results and Analyses}

We additionally provide an inference latency analysis of FactorDrive on NAVSIM to assess its inference efficiency. To further illustrate FactorDrive's adaptive multi-step reasoning and trajectory planning capabilities, we present qualitative examples with different planning demands on nuScenes and NAVSIM. The green and red trajectories denote the predicted and ground-truth trajectories, respectively. Alongside the final trajectory visualization, we present the model-generated reasoning path, meta-action, and future ego trajectory to analyze how different planning-critical factors (PCFs) affect reasoning-path composition and depth.

\subsection{Inference Latency Analysis}
\begin{figure}[!t]
    \centering
    \includegraphics[width=0.95\columnwidth]{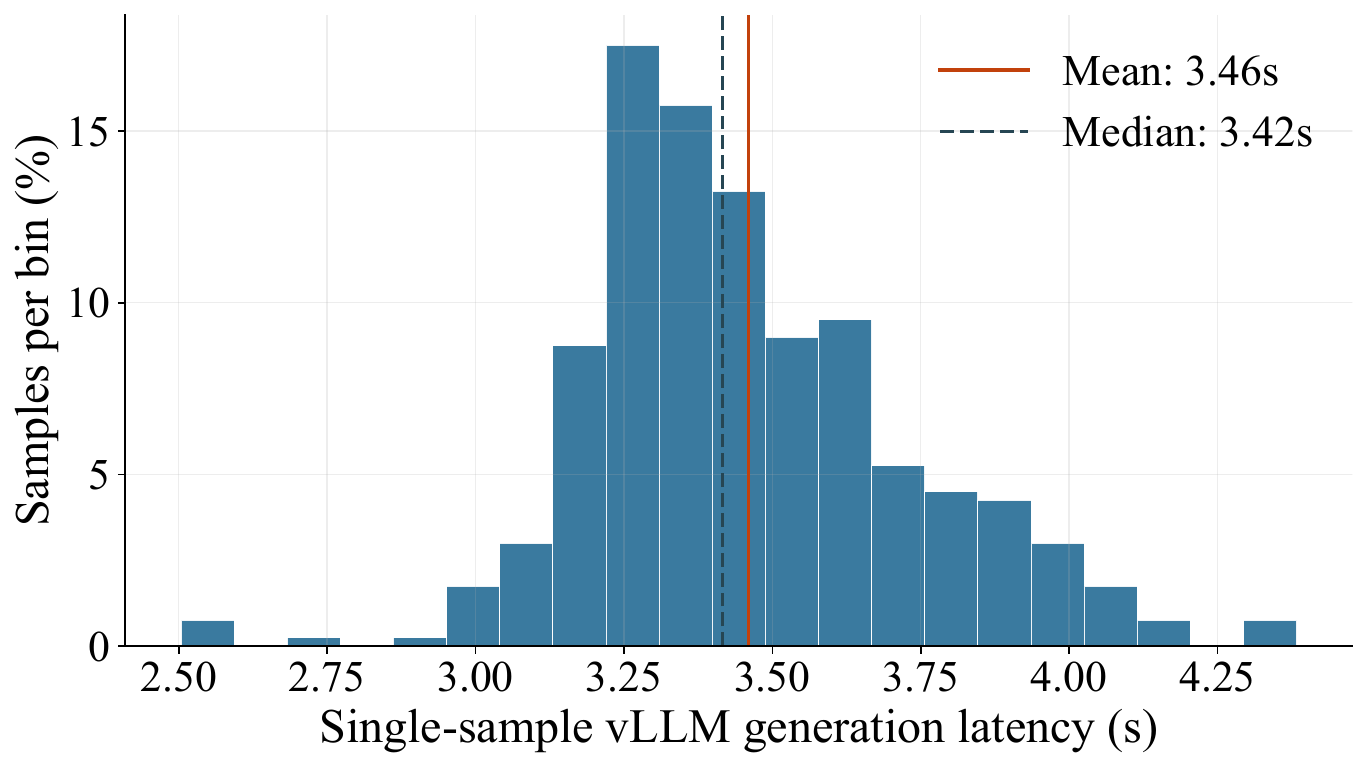}
    \caption{Distribution of single-sample inference latency on NAVSIM.}
    \label{fig:navsim_inference_time_400_valid}
\end{figure}

To evaluate the inference efficiency of FactorDrive, we randomly sample 400 examples from the NAVSIM \texttt{navtest} split and measure the per-sample inference latency with a batch size of 1. All experiments are conducted using vLLM 0.11.0~\cite{kwon2023vllm} on a single NVIDIA H100 GPU with BF16 precision. As shown in Fig.~\ref{fig:navsim_inference_time_400_valid}, the mean and median inference latencies are \(3.46\,\mathrm{s}\) and \(3.42\,\mathrm{s}\), respectively. The observed latency variation may be partly attributed to differences in generated sequence lengths across driving scenarios. For practical deployment, low-precision quantization techniques can be adopted to further reduce computational and memory overhead and improve inference efficiency~\cite{xiao2023smoothquant,lin2024awq}.

\subsection{Qualitative Results on nuScenes}

\begin{figure*}[t]
    \centering
    \includegraphics[width=0.98\textwidth]{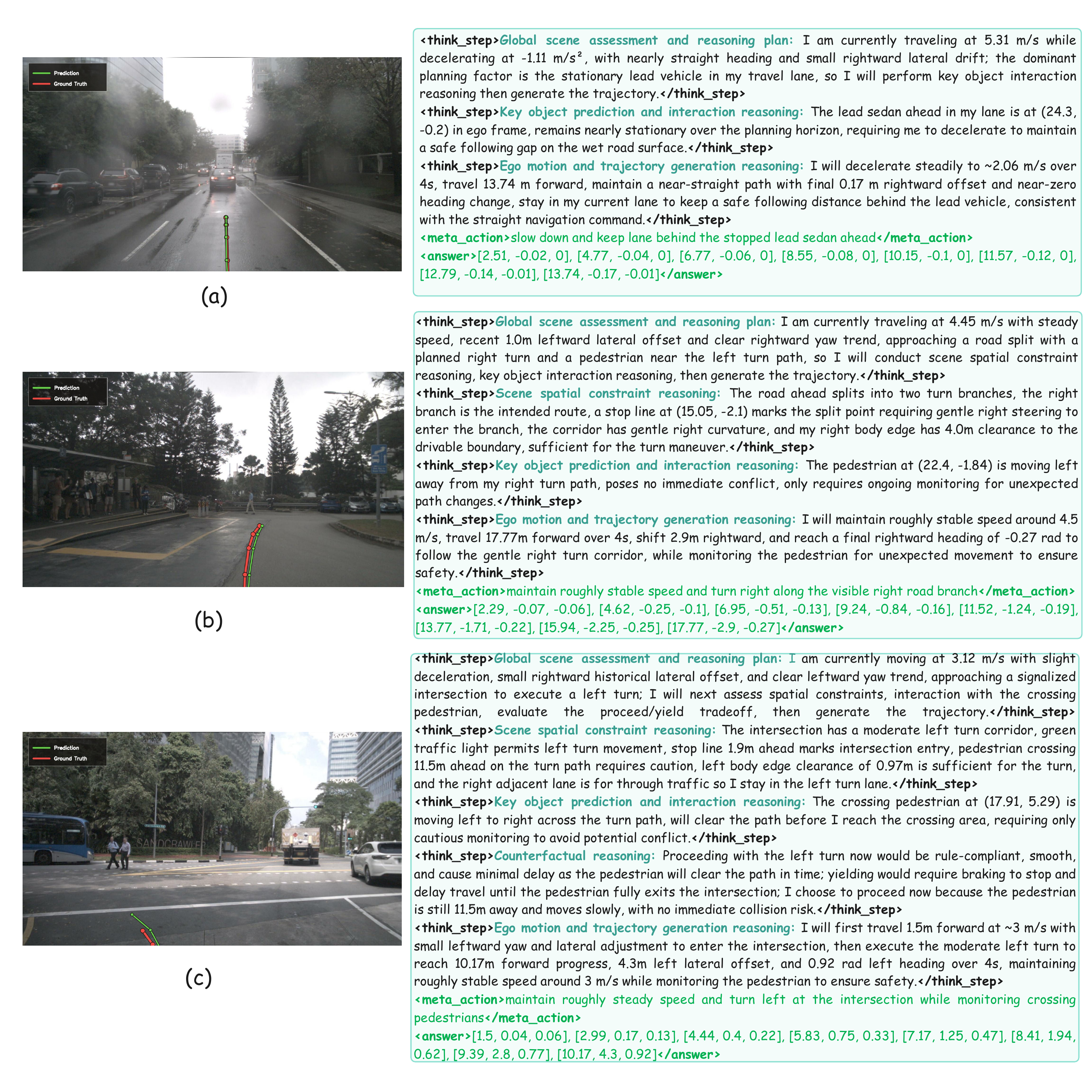}
    \caption{Qualitative results on nuScenes. Green and red denote the predicted and ground-truth trajectories, respectively. FactorDrive adapts reasoning-path composition and depth to the planning-critical factors active in each scene.}
    \label{fig:app_nuscenes_qualitative}
\end{figure*}

Fig.~\ref{fig:app_nuscenes_qualitative} presents three nuScenes scenarios with different planning demands. As the number of active PCFs and decision constraints increases, FactorDrive does not follow a uniform reasoning structure but instead selects the reasoning units relevant to the current scene.

In Fig.~\ref{fig:app_nuscenes_qualitative}(a), the ego vehicle approaches a nearly stationary lead vehicle in the same lane. FactorDrive identifies the lead vehicle as the dominant planning factor through Global Scene Assessment and Reasoning Plan. It then activates Key Object Prediction and Interaction Reasoning and uses the relative motion of the lead vehicle to determine that continued deceleration is required. In Ego Motion and Trajectory Generation Reasoning, the model generates a near-straight trajectory that slows the ego vehicle while maintaining its lane and a safe following gap. Because no additional spatial-corridor constraint or genuine action trade-off affects the current plan, the reasoning path does not include Scene Spatial Constraint Reasoning or Counterfactual Reasoning. This three-step reasoning path shows that FactorDrive organizes the necessary analysis around the active PCFs rather than applying a fixed structure or exhaustively describing all visible elements.

Fig.~\ref{fig:app_nuscenes_qualitative}(b) presents a right-turn scenario at a road split. Compared with Fig.~\ref{fig:app_nuscenes_qualitative}(a), this scene simultaneously involves turn-corridor geometry, a drivable-area boundary, and a nearby pedestrian. FactorDrive therefore activates Scene Spatial Constraint Reasoning and Key Object Prediction and Interaction Reasoning after Global Scene Assessment and Reasoning Plan. The former identifies the feasible right-turn corridor and lateral clearance, whereas the latter determines that the pedestrian is moving away from the planned path and therefore requires continued monitoring rather than an immediate stop. The model ultimately generates a trajectory that follows the feasible road branch and completes the right turn smoothly. This four-step reasoning path shows that FactorDrive introduces the required reasoning units when road geometry and agent interaction jointly affect planning.

Fig.~\ref{fig:app_nuscenes_qualitative}(c) further presents a left-turn scenario at a signalized intersection. In addition to the traffic-light constraint, turn-corridor geometry, and crossing pedestrian, the scene involves a genuine proceed-versus-yield decision trade-off. FactorDrive therefore uses the complete five-step reasoning path and further activates Counterfactual Reasoning after Spatial Reasoning and Interaction Reasoning. The model compares the effects of turning left immediately and stopping to wait on safety, driving efficiency, and trajectory shape. Based on the green-light state, the pedestrian's relative motion, and the temporal relationship between the pedestrian and the future ego corridor, it chooses to proceed with the left turn. The final predicted trajectory remains close to the ground-truth trajectory while reflecting the corresponding lateral displacement and heading change.

These three scenes present three-step, four-step, and five-step reasoning paths, respectively. Their differences are not simply determined by the number of visible objects but by the PCFs that actively participate in the current planning decision. These examples intuitively illustrate FactorDrive's core design: reasoning-path composition and depth vary with scene-specific planning demands, thereby avoiding insufficient analysis in constrained scenes and unnecessary Counterfactual Reasoning when no decision trade-off exists.

\subsection{Qualitative Results on NAVSIM}

\begin{figure*}[t]
    \centering
    \includegraphics[width=0.98\textwidth]{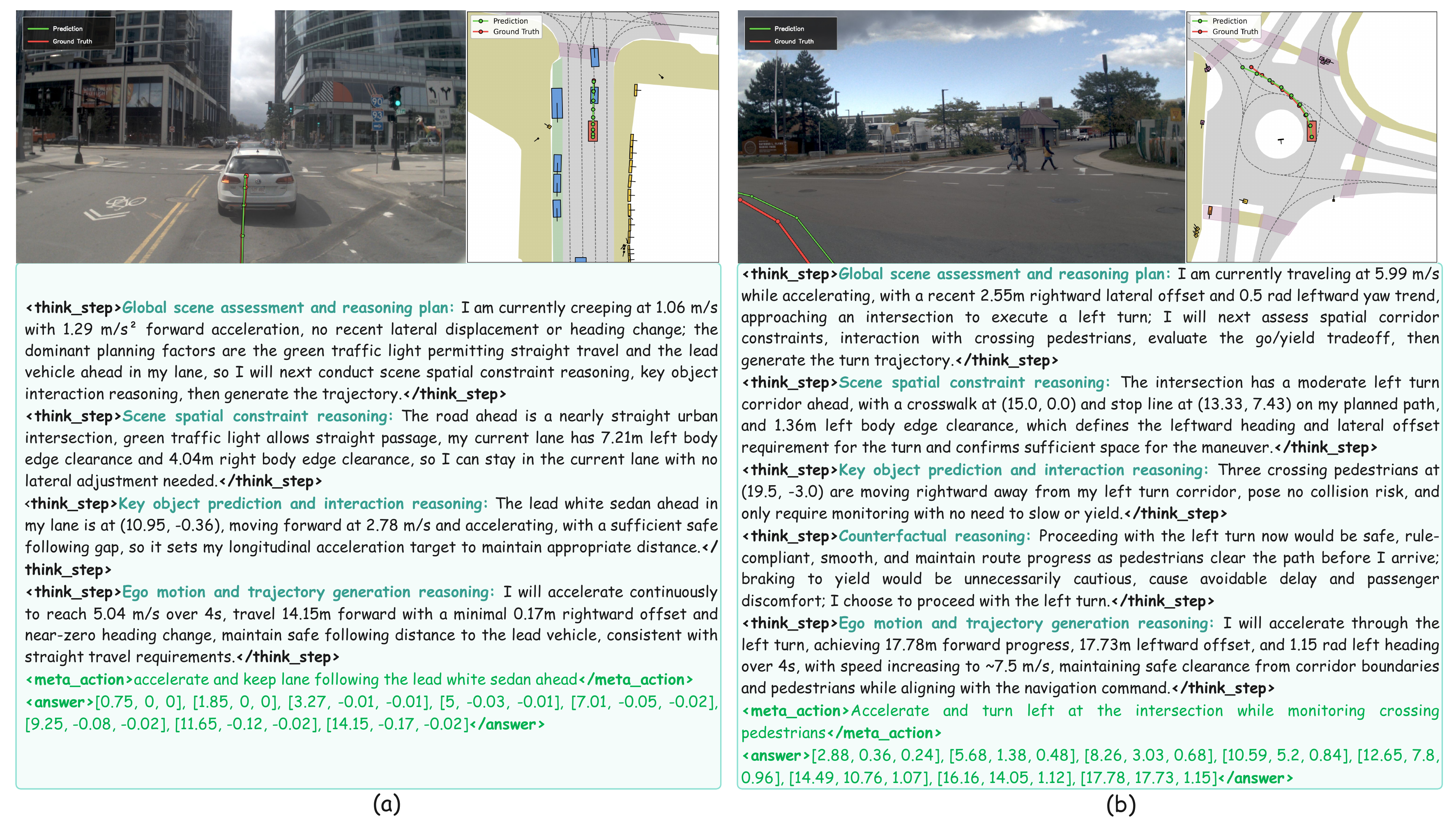}
    \caption{Qualitative results on NAVSIM. FactorDrive selects scene-relevant reasoning units to integrate traffic constraints, road geometry, and critical-agent interactions into trajectory generation.}
    \label{fig:app_navsim_qualitative}
\end{figure*}

Fig.~\ref{fig:app_navsim_qualitative} presents two NAVSIM scenarios with different planning demands to illustrate how FactorDrive transforms spatial-physical evidence into planning reasoning that directly supports trajectory generation.

Fig.~\ref{fig:app_navsim_qualitative}(a) presents a relatively straightforward straight-through following scenario, although the current planning decision is still jointly constrained by the green traffic light and lead vehicle. FactorDrive uses Scene Spatial Constraint Reasoning to confirm that the traffic signal permits straight travel and that the current lane provides sufficient lateral clearance. It then uses Key Object Prediction and Interaction Reasoning to analyze the relative position, velocity, and motion trend of the lead vehicle. Because no reasonable alternative action exists, the model does not activate Counterfactual Reasoning. Instead, in Motion Reasoning, it generates a trajectory that gradually accelerates, remains within the current lane, and maintains a safe following distance. The predicted trajectory largely overlaps the ground-truth trajectory, indicating that the model can jointly incorporate the traffic-control constraint and agent motion into longitudinal planning.

Fig.~\ref{fig:app_navsim_qualitative}(b) presents a left-turn scenario involving crossing pedestrians. FactorDrive first identifies spatial constraints such as the left-turn corridor, crosswalk, and stop line, and then determines that the three pedestrians are moving rightward and gradually leaving the future ego corridor. Because the scene involves a decision trade-off between proceeding with the turn and decelerating to yield, the model further activates Counterfactual Reasoning. It compares the effects of proceeding and yielding on safety, route progress, and comfort, and ultimately chooses to complete the left turn while continuing to monitor the pedestrians. The generated turning trajectory remains well aligned with the ground-truth trajectory and satisfies the constraints imposed by the road boundaries and pedestrian interactions.

Both NAVSIM scenarios require spatial and interaction reasoning, but Counterfactual Reasoning is activated only in Fig.~\ref{fig:app_navsim_qualitative}(b), where a genuine decision trade-off exists. This comparison further shows that FactorDrive's adaptation is not merely a switch between generating CoT and not generating CoT. Instead, it selects specific reasoning units according to the active PCFs within a unified reasoning framework.

\subsection{Qualitative Comparison Before and After QS-GRPO}

\begin{figure*}[t]
    \centering
    \includegraphics[width=0.98\textwidth]{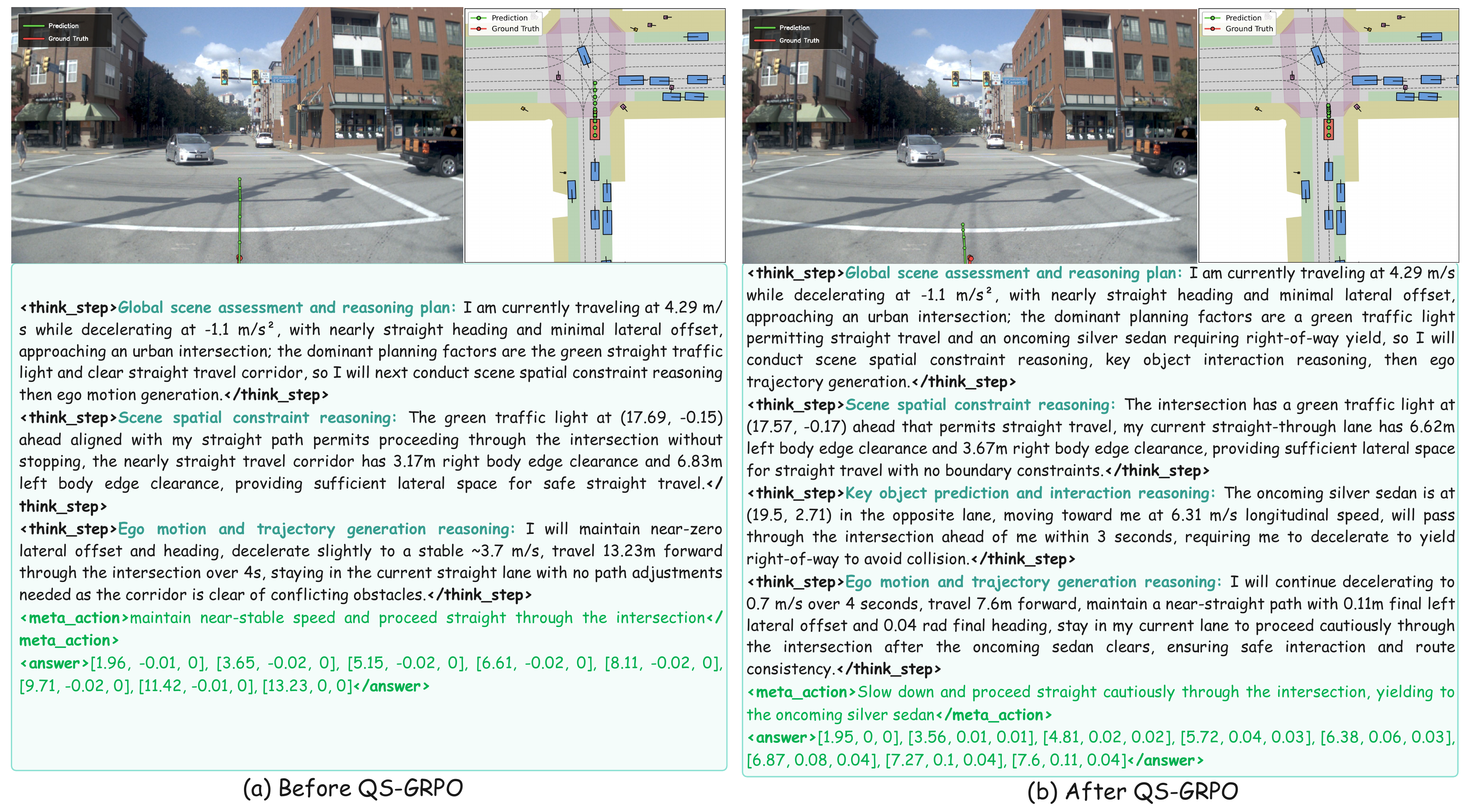}
    \caption{Qualitative comparison of the PCF-CoT SFT model before and after QS-GRPO post-training. QS-GRPO enables the model to identify the previously overlooked oncoming vehicle, introduce the required interaction reasoning, and generate a trajectory that more closely follows the ground truth.}
    \label{fig:app_qs_grpo_qualitative}
\end{figure*}

Fig.~\ref{fig:app_qs_grpo_qualitative} compares the reasoning and planning outputs of the PCF-CoT SFT model on the same NAVSIM scenario before and after QS-GRPO post-training.

Before QS-GRPO, the model correctly identifies the green traffic light and clear straight-through corridor but fails to recognize the oncoming silver sedan as a critical interacting agent that affects the current right-of-way decision. Its reasoning path therefore contains only Global Scene Assessment and Reasoning Plan, Scene Spatial Constraint Reasoning, and Ego Motion and Trajectory Generation Reasoning. The model chooses to proceed directly through the intersection at a near-stable speed and predicts approximately $13.23\,\mathrm{m}$ of forward progress, substantially exceeding the safe progress indicated by the ground-truth trajectory.

After QS-GRPO post-training, the model includes the oncoming silver sedan among the dominant planning factors and additionally activates Key Object Prediction and Interaction Reasoning. Based on the vehicle's ego-centric position and relative motion, the model determines that it will pass through the conflict area first and therefore changes the meta-action from proceeding directly to slowing down and yielding cautiously. Accordingly, the predicted forward progress decreases from $13.23\,\mathrm{m}$ to $7.60\,\mathrm{m}$, and the trajectory endpoint becomes closer to the ground truth.

This case shows that the changes introduced by QS-GRPO are reflected not only in the final trajectory but also in reasoning-path composition and the PCFs on which the reasoning is based. The post-trained model supplements the Interaction Reasoning omitted by the original model and transforms this information into more appropriate longitudinal planning. This behavior is consistent with the design objective of QS-GRPO: trajectory-level planning rewards guide quality-guided MCTS to explore reasoning paths that yield higher planning quality, and group-relative policy optimization reinforces these paths.

\section{Ethical Statement}
FactorDrive is intended solely for research on end-to-end autonomous driving. We use publicly available datasets under their respective licenses, collect no new real-world driving data, and do not annotate or infer identity-related information; human participation is limited to quality inspection of generated annotations rather than human-subject experiments. Because PCF-CoT annotations are generated by Seed-2.0-Pro, errors or biases may remain despite automatic validation and human inspection. Moreover, results on the open-loop nuScenes and closed-loop-oriented NAVSIM benchmarks do not establish safety in real-world driving. FactorDrive should therefore not be deployed as a standalone driving system without extensive closed-loop and on-road validation and appropriate safety mechanisms.

\section{Reproducibility Statement}
The main paper and appendix provide detailed descriptions of the PCF-CoT data construction and quality-control pipeline, the two-stage supervised fine-tuning and QS-GRPO post-training procedures, as well as the evaluation protocols, reward formulations, and training configurations. Upon acceptance, we will release the PCF-CoT dataset, source code, and model checkpoints through a public project repository. Source datasets and third-party models remain subject to their respective licenses. For resources that cannot be redistributed, we will provide detailed instructions for obtaining them from their original sources.

\end{document}